\documentclass[dvipsnames]{article} %
\usepackage{colm2024_conference}

\usepackage{booktabs}
\usepackage{enumitem}
\usepackage{wrapfig}
\usepackage{algorithm}
\usepackage{algpseudocode}
\usepackage{graphicx}
\usepackage{subfigure}
\usepackage[misc]{ifsym}

\usepackage{microtype}
\usepackage{amsmath}
\usepackage{colortbl}
\usepackage[utf8]{inputenc}
\usepackage[T1]{fontenc}
\definecolor{lightgray}{rgb}{0.9,0.9,0.9}
\usepackage{caption}
\usepackage{subcaption}
\usepackage{setspace}
\usepackage{url}
\usepackage{multirow}
\usepackage{tabularx}
\usepackage{blindtext}
\usepackage{pgfplots}
\pgfplotsset{compat=1.18} 
\usepackage{tikz}
\usetikzlibrary{er,positioning,bayesnet}
\usepackage{makecell}
\usepackage{tipa}
\usepackage{siunitx}
\usepackage{nicefrac}
\usepackage{listings}
\usepackage[raster,skins, most]{tcolorbox} %
\usepackage{xltabular}
\usepackage{adjustbox}
\usepackage{xurl}
\usepackage{rotating}
\usepackage[normalem]{ulem}
\usepackage{xcolor}
\newcommand{\symboleeasysingle}{\raisebox{0pt}{\includegraphics[scale=0.25]{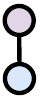}}~}
\newcommand{\symboleeasymuti}{\raisebox{0pt}{\includegraphics[scale=0.25]{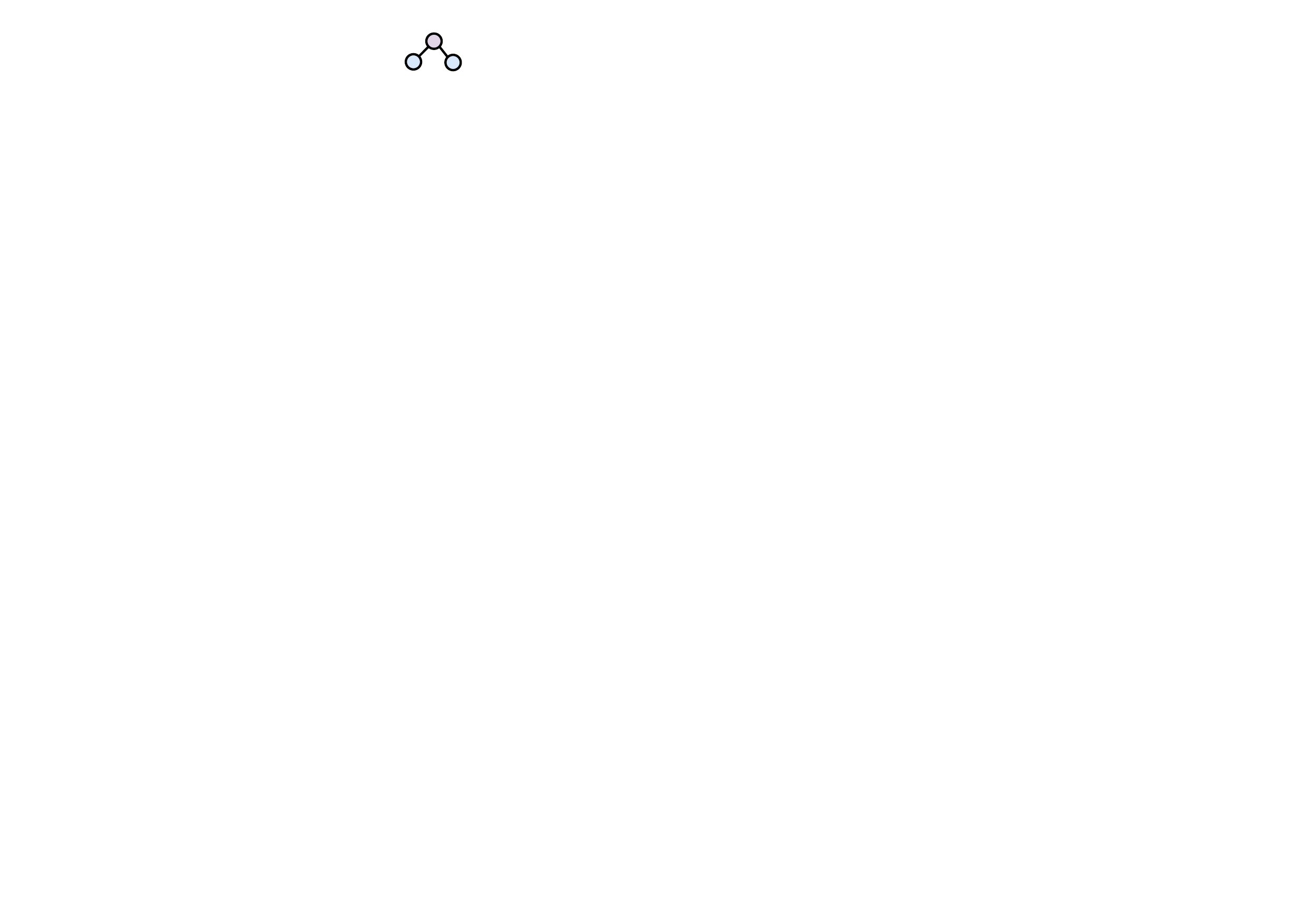}}~}
\newcommand{\symbolmediumsingle}{\raisebox{0pt}{\includegraphics[scale=0.25]{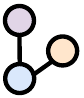}}~}
\newcommand{\symbolmediummuti}{\raisebox{0pt}{\includegraphics[scale=0.25]{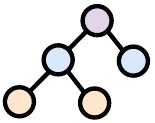}}~}
\newcommand{\symbolehardsingle}{\raisebox{0pt}{\includegraphics[scale=0.25]{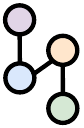}}~}
\newcommand{\symbolehardmuti}{\raisebox{0pt}{\includegraphics[scale=0.25]{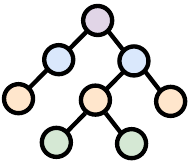}}~}
\usepackage{fontawesome}

\useunder{\uline}{\ul}{}

\usepackage{amsmath,amsfonts,bm}

\usepackage{amssymb}

\definecolor{babyred}{rgb}{0.85, 0.93, 0.97}

\definecolor{babygreen}{rgb}{0.85, 0.97, 0.85}

\definecolor{mycell}{gray}{.95}

\def\eqref#1{equation~\ref{#1}}

\def\1{\bm{1}}

\DeclareMathAlphabet{\mathsfit}{\encodingdefault}{\sfdefault}{m}{sl}
\SetMathAlphabet{\mathsfit}{bold}{\encodingdefault}{\sfdefault}{bx}{n}

\newtcolorbox{findings}[1][]{
  	title=#1,
        colframe=uclablue,
        top=2pt,           
        bottom=0pt,        
        left=0pt,          
        right=0pt,          
        before skip=0pt,        
        after skip=0pt,          
}
\newcommand{\hfdataset}{\raisebox{-1.5pt}{\includegraphics[height=1.05em]{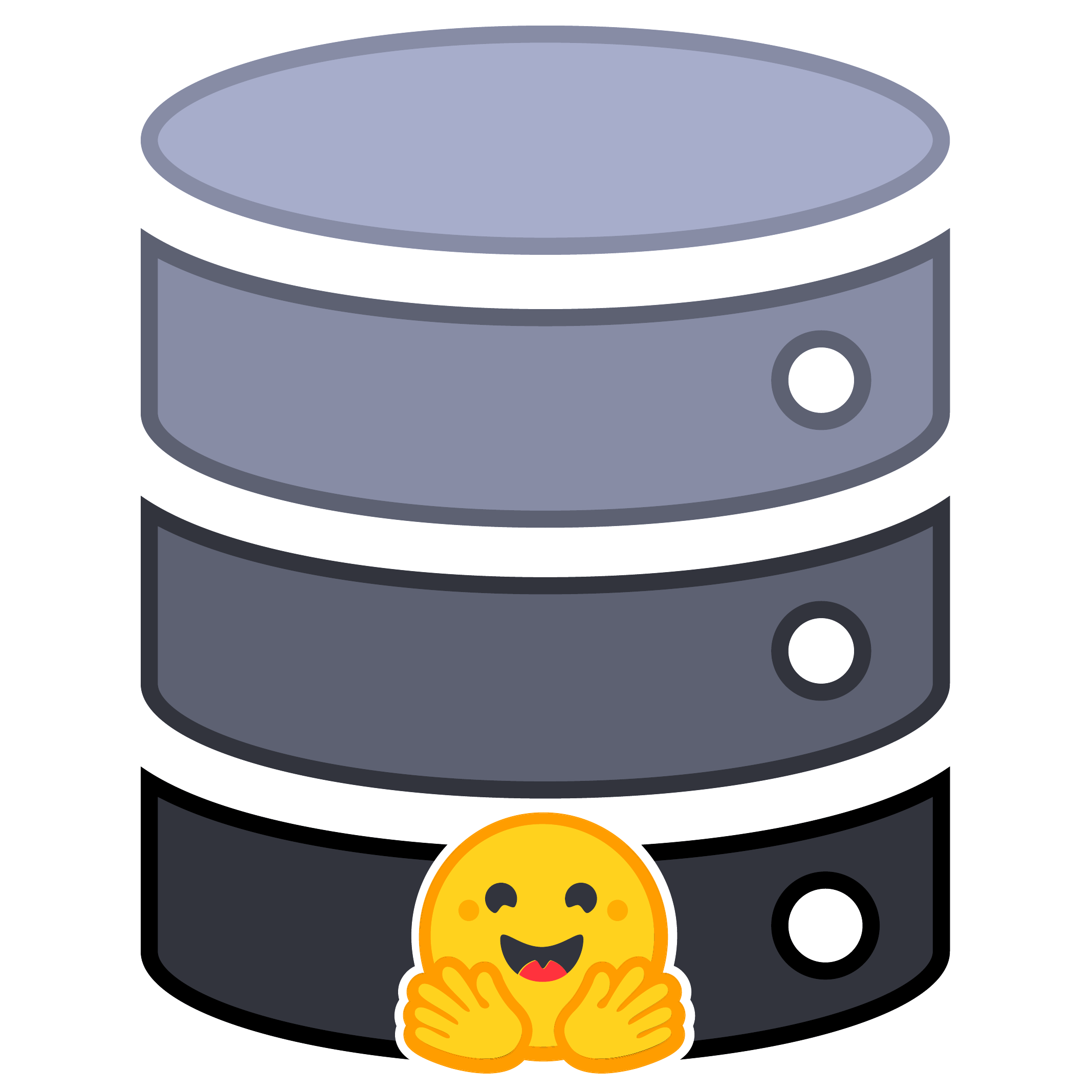}}\xspace}
\definecolor{royalblue(web)}{rgb}{0.25, 0.41, 0.88}
\definecolor{whitesmoke}{rgb}{0.96, 0.96, 0.96}
\definecolor{codegreen}{rgb}{0,0.6,0}
\definecolor{codegray}{rgb}{0.5,0.5,0.5}
\definecolor{codepink}{RGB}{252, 142, 172}
\definecolor{codepurple}{rgb}{0.58,0,0.82}
\lstdefinestyle{mystyle}{
    language=Python,
    commentstyle=\color{codegreen},
    keywordstyle=\color{magenta},
    numberstyle=\tiny\color{codegray},
    stringstyle=\color{codepurple},
    basicstyle=\ttfamily \lst@ifdisplaystyle\tiny\fi,
    breakatwhitespace=false,
    breaklines=true,
    captionpos=b,
    keepspaces=true,
    numbers=left,
    numbersep=5pt,
    xleftmargin=0pt,  
    showspaces=false,  
    showstringspaces=false,  
    showtabs=false,
    tabsize=2,
    columns=flexible,  
    moredelim=[is][\bfseries]{<highlight>}{</highlight>}
}

\newtcolorbox{json}[1][]{
	float,
  	title=#1,
        top=1pt,           
        bottom=1pt,        
        left=0pt,          
        right=0pt,          
        before skip=0.65em, after skip=0.75em,
}

\newcommand{\symbolecritic}{\raisebox{0pt}{\includegraphics[scale=0.23]{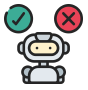}}~}
\newcommand{\symbolexplorer}{\raisebox{0pt}{\includegraphics[scale=0.23]{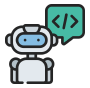}}~}

\renewcommand{\ghlink}{https://github.com/Alibaba-NLP/WebAgent}

\newcommand*\justify{%
  \fontdimen2\font=0.4em
  \fontdimen3\font=0.2em
  \fontdimen4\font=0.1em
  \fontdimen7\font=0.1em
  \hyphenchar\font=`\-
}

\renewcommand{\texttt}[1]{%
  \begingroup
  \ttfamily
  \begingroup\lccode`~=`/\lowercase{\endgroup\def~}{/\discretionary{}{}{}}%
  \begingroup\lccode`~=`[\lowercase{\endgroup\def~}{[\discretionary{}{}{}}%
  \begingroup\lccode`~=`.\lowercase{\endgroup\def~}{.\discretionary{}{}{}}%
  \catcode`/=\active\catcode`[=\active\catcode`.=\active
  \justify\scantokens{#1\noexpand}%
  \endgroup
}

\usepackage{makecell}
\usetikzlibrary{tikzmark}
\makeatletter
\newcommand*\myfontsize{%
  \@setfontsize\myfontsize{7}{8}%
}
\makeatother

\definecolor{uclablue}{RGB}{159, 195, 224}

\definecolor{uclagold}{RGB}{255, 240, 180}

\definecolor{aliceblue}{RGB}{255, 238, 241}

\definecolor{cadmiumgreen}{rgb}{0.0, 0.42, 0.24}

\definecolor{myred}{rgb}{0.7, 0.3, 0.0}
\definecolor{myblue}{rgb}{0.2, 0.3, 0.6}
\definecolor{babygreen}{rgb}{0.85, 0.97, 0.85}

\definecolor{purple1}{RGB}{126, 107, 196}
\definecolor{purple2}{RGB}{199, 158, 207}
\definecolor{purple3}{RGB}{214, 200, 255}
\definecolor{purple4}{RGB}{254, 240, 255}

\definecolor{deepblue}{RGB}{48, 58, 82}

\newcommand{\symboletongyi}{\raisebox{0pt}{~\includegraphics[scale=0.012]{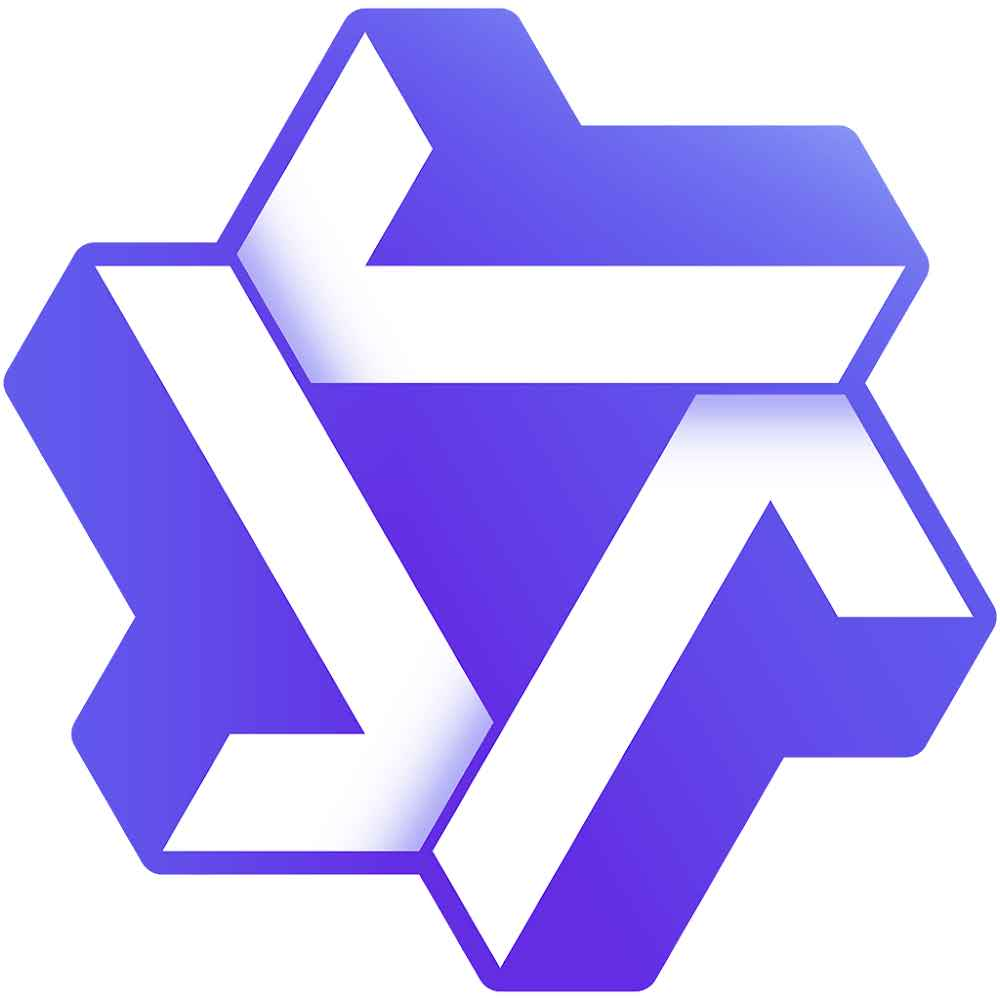}}~}

\definecolor{deepPurple}{HTML}{330066}

\definecolor{uclablue_old}{rgb}{0.15, 0.45, 0.68}
\hypersetup{
    breaklinks,
    citecolor=uclablue_old,
    colorlinks=true,
}

\newtcolorbox{mybox}[2][]
  {colback = black!5!white, colframe = black!75!black, fonttitle = \bfseries,
    colbacktitle = black!100!black, enhanced, before upper={\fontsize{8}{11}\obeyspaces\obeylines\selectfont}, fontupper=\selectfont,
    attach boxed title to top left={yshift=-2.2mm,xshift=4mm},
    title=#2,#1}

\title{WebWalker: Benchmarking LLMs in Web Traversal}

\author{%
Jialong Wu, Wenbiao Yin, Yong Jiang$^{(\textrm{\Letter})}$, Zhenglin Wang, Zekun Xi, Runnan Fang, Linhai Zhang, Yulan He, Deyu Zhou, Pengjun Xie, Fei Huang%
  \\[1em]               
  {\fontsize{10pt}{11pt}\selectfont          
Tongyi Lab\symboletongyi, Alibaba Group}\\
}

\newcommand{\eg}{\textit{e.g.}}

\newcommand{\symbolclick}{\raisebox{0pt}{\includegraphics[scale=0.05]{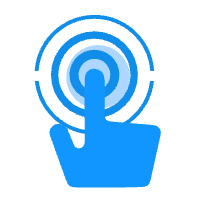}}~}
\usepackage{pifont}
\newcommand{\yessymbol}{\color{green!60!black}\ding{51}}
\newcommand{\nosymbol}{\color{red!60!black}\ding{55}}
\begin{document}

\maketitle

\begingroup
  \renewcommand\thefootnote{\Letter}  
  \footnotetext{Corresponding author.} 
\endgroup

\begin{abstract}
Retrieval-augmented generation (RAG) demonstrates remarkable performance across tasks in open-domain question-answering.
However, traditional search engines may retrieve shallow content, limiting the ability of LLMs to handle complex, multi-layered information.
To address it, we introduce \textbf{WebWalkerQA}, a benchmark designed to assess the ability of LLMs to perform web traversal.
It evaluates the capacity of LLMs to traverse a website's subpages to extract high-quality data systematically. 
We propose \textbf{WebWalker}, which is a multi-agent framework that mimics human-like web navigation through an explore-critic paradigm.
Extensive experimental results show that WebWalkerQA is challenging and demonstrates the effectiveness of RAG combined with WebWalker, through the horizontal and vertical integration in real-world scenarios.
\end{abstract}


\section{Introduction}
\label{sec:intro}

Large Language Models (LLMs) have demonstrated impressive capabilities across a wide range of natural language processing tasks~\citep{ouyang2022training,chatgpt}.
While their knowledge base remains static post-training, integrating external search engines via retrieval-augmented generation (RAG) allows LLMs to retrieve up-to-date information from the web, enhancing their utility in dynamic, knowledge-intensive scenarios~\citep{rag}.
However, traditional online search engines, \eg, Google or Bing, perform horizontal searches of queries and may not effectively trace the deeper content embedded within websites.

Interacting with the web pages and digging through them can effectively address this issue.
Previous works related to web pages focus on addressing action-based requests, such as Mind2Web~\citep{deng2023mindweb} and WebArena~\citep{webarena}; these HTML-based instruction-action benchmarks face challenges such as excessively noisy information and overly long inputs, which can significantly hinder performance due to limitations in long-context understanding. 
Additionally, they fail to capture the complexities of real-world scenarios where relevant information is buried deep within web pages and requires multiple layers of interaction.

To fill this gap, a new task \textbf{Web Traversal} is proposed, given an initial website corresponding to a query, systematically traverses web pages to uncover information.
We propose \textbf{WebWalkerQA}, designed specifically to evaluate LLMs on their ability to handle queries embedded in complex, multi-step web interactions on a given root website.
WebWalkerQA focuses on text-based reasoning abilities, using a Question-Answer format to evaluate traversal and problem-solving capabilities in web scenarios.
We constrain actions to ``\textit{click}'' \symbolclick to evaluate the agent's navigation and information-seeking capabilities.
This paradigm is more targeted and aligns better with practical applications.
WebWalkerQA reflects real-world challenges, emphasizing the \textbf{\textit{\underline{depth}}} of the source information across education, conference, organization, and game domains,  where official sources are published and paths to information are more structured with clickable buttons and reasoning logic.
Several types, including \texttt{multi-source} and \texttt{single-source} QAs, are developed to evaluate the ability of LLMs to mimic different human web-navigation paradigms. 

Additionally, we introduce a strong baseline \textbf{WebWalker}, a multi-agent framework designed to emulate human-like web navigation through vertical exploration.
The framework consists of an explorer agent and a critic agent.
Given the need for reasoning capabilities to navigate and interact with web pages effectively, the explorer agent is built upon the ReAct framework~\citep{react}, 
\begin{wrapfigure}{r}{8cm}
\centering\includegraphics[width=0.45\textwidth]{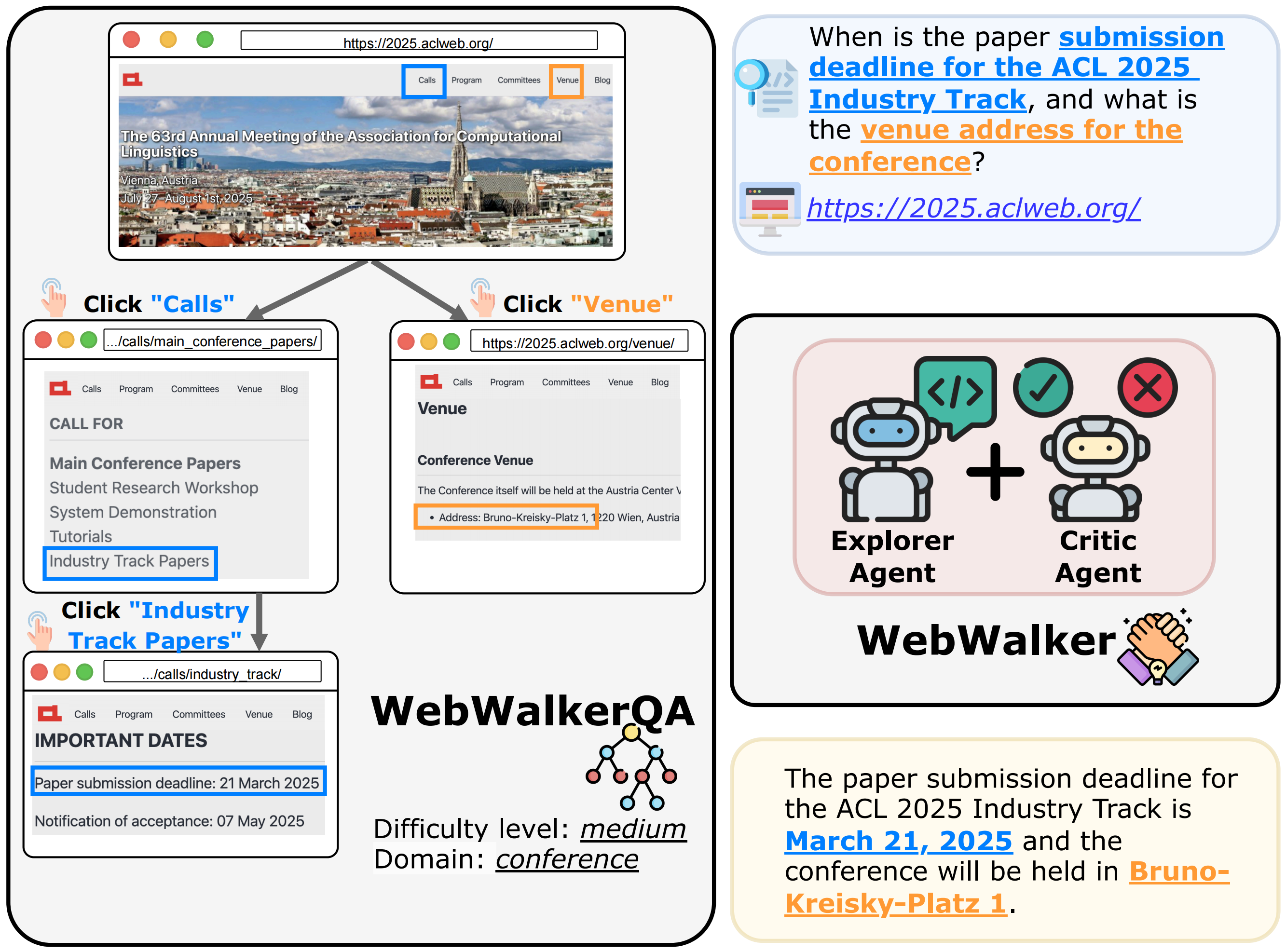}
    \caption{A \texttt{multi-source} QA\protect\footnotemark example from WebWalkerQA that requires traversing web pages to gather information for answering the given question.}
    \label{fig:intro}
\end{wrapfigure}
\footnotetext{In our paper, \texttt{multi-source} refers to the requirement of information from multi distinct web pages.}
leveraging a thought-action-observation paradigm, while the critic agent is responsible for maintaining memory and generating responses based on the exploration conducted by the explorer agent.


We evaluate the performance of the WebWalker, built on various mainstream LLMs, including both closed-source and open-sourced, using WebWalkerQA as the benchmark. 
However, even with the most powerful LLMs as the backbone, its performance on WebWalkerQA remains suboptimal, thereby validating the challenge posed by WebWalkerQA.

We then conduct further experiments to validate the integration with the RAG for information-seeking QA tasks.
Our findings are as follows:
\textbf{\textit{(i)}}  Web navigation still requires efforts in tasks that demand planning and reasoning;
\textbf{\textit{(ii)}} By combining RAG with the WebWalker, this horizontal and vertical coordination proves effective;
\textbf{\textit{(iii)}} Vertical exploration of pages offers a promising direction for scaling inference time in RAG systems.

The contributions of our work are as follows:
\begin{itemize}[itemsep=0.5pt, parsep=0pt]
\vspace{-2mm}
    \item We construct a challenging benchmark, \textbf{WebWalkerQA}, which is composed of 680 queries from four real-world scenarios across over 1373 webpages.
    \item To tackle the challenge of web-navigation tasks requiring long context, we propose \textbf{WebWalker}, which utilizes a multi-agent framework for effective memory management.
    \item Extensive experiments show that the WebWalkerQA is challenging, and for information-seeking tasks, vertical exploration within the page proves to be beneficial.
\end{itemize}
\section{Related Work}

\subsection{Web-Oriented Benchmark}
Before the era of LLMs, several web-oriented benchmarks had already been proposed~\citep{liu2018reinforcement,xu-etal-2021-grounding,humphreys2022data,yao2022webshop,mialon2024gaia,xu2024turkingbenchchallengebenchmarkweb}.
LLMs are capable of interacting with complex environments, like the open web in HTML or DOM format~\citep{tan2024htmlrag}, leading to the development of an increasing number of benchmarks aimed at evaluating the interaction capabilities of LLMs with web content.
The widely used benchmark today, Mind2Web~\citep{deng2023mindweb}, is a dataset designed for evaluating web agents that follow instructions to complete complex tasks, typically through multiple-choice questions.
Subsequent works have extended the interaction to the vision domain,  incorporating information from screenshots~\citep{seeact,zheng-etal-2024-webolympus,he-etal-2024-webvoyager,koh2024visualwebarena,cheng2024seeclick}.
The web-oriented benchmark is becoming progressively more human-like, vision-centric, and increasingly broad, complex, and realistic~\citep{liuagentbench,hong2024cogagent,kim2024language,zhang2024mmina}.
The most closely to ours are the MMInA~\citep{zhang2024mmina} and AssistantBench~\citep{assistantbench}, both of which focus on time-consuming tasks that require navigation across multiple pages.
In our work, WebWalkerQA takes the form of QA pairs. 
Unlike all previous works, we construct both single-source and multi-source queries from the width perspectives of the website, aiming to simulate two types of page exploration patterns typically exhibited by humans.
The comparison between WebWalkerQA and other benchmarks is shown in Table~\ref{table:comparison}.

\begin{table*}[]
\small
\centering
\begin{tabular}{@{}c|c|c|c|c|c|c@{}}
\toprule
 & \textbf{Language}  & \textbf{Format} & \textbf{Depth} & \textbf{Width} & \textbf{Hop} & \textbf{\# Pages}
\\
\toprule
Mind2Web~\citep{deng2023mindweb}  & En & Multi-choice & \nosymbol & \nosymbol & \nosymbol & 100
\\
WebArena~\citep{webarena}  & En & Action & \nosymbol & \nosymbol & \nosymbol & 6
\\
AssistantBench~\citep{assistantbench} & En & QA &  \nosymbol & \yessymbol &\yessymbol & 525 \\
MMInA~\citep{zhang2024mmina} &En&Action& \nosymbol & \yessymbol & \yessymbol &100\\
GAIA~\citep{mialon2024gaia} &En&QA& \nosymbol & \yessymbol & \yessymbol & -\\
\midrule
\textbf{WebWalkerQA}  & En\&Zh & QA &  \yessymbol & \yessymbol & \yessymbol &1373\\
\bottomrule 
\end{tabular}
\caption{Comparison between WebWalkerQA and other benchmarks.
\textbf{Depth} refers to the extent of exploration required on a given website. 
\textbf{Width} denotes whether answering a query necessitates multiple sources.
\textbf{Hop} indicates whether multiple steps are required to complete the task. 
\textbf{\#Pages} refers to the number of webpages involved.}
\label{table:dataset_type}
\end{table*}
\subsection{Agents on Web-Navigation}
Based on web-oriented benchmarks,  numerous web agents have been proposed~\citep{nakano2021webgpt,liu2023webglm,zhou2023agents,lai2024autowebglm,zhou2024symbolic}.
Web agents primarily follow two lines of development: one leverages a small language model trained specifically to filter actions or identify relevant HTML elements~\citep{seeact,mt_mind2web_agent,furuta2024multimodal}.
The other line focuses on prompting LLMs
~\citep{reddy2024infogent,song2024beyond,treesearchagent}, where different agentic modules are used to guide the model in accomplishing complex web navigation tasks more effectively. 
In addition, with the rise of visual web-oriented benchmarks, many agents now use screenshots as sensory input~\citep{he2024openwebvoyager,abuelsaad2024agent,iong2024openwebagent}.
Unlike previous works, WebWalker specializes in information-seeking by reasoning over HTML button data.
It emulates human-like page interactions with web pages to access reliable, authoritative information utilizing a multi-agent framework.

\section{WebWalkerQA}
\begin{figure*}[htbp]
 \centering
 \includegraphics[width=1.0\textwidth]{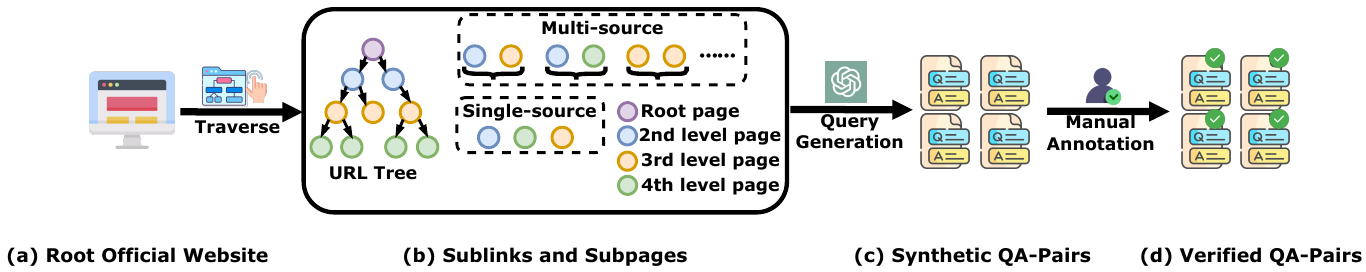}
 \caption{Data Generation Pipeline for \textbf{WebWalkerQA}. We first collect root official websites across conference, organization, education, and game domains.
 Then we mimic human behavior by systematically clicking and collecting subpages accessible through sublinks on the root page. 
 Using predefined rules, we leverage GPT4o to generate synthetic QA-pairs based on the gathered information, followed by manual verification to ensure accuracy and relevance.
 }
\label{fig:annoatation}
\end{figure*}

We present WebWalkerQA in this section, starting with an overview of the data collection process to ensure quality (\S\ref{sec:data_collection}), followed by a discussion of WebWalkerQA's statistics (\S\ref{sec:data_statistics}).
Finally, we introduce the new task, Web Traversal, and describe the evaluation metrics for WebWalkerQA(\S\ref{sec:evaluation}).

\subsection{Data Collection}
\label{sec:data_collection}
To make the annotation process cost-efficient and accurate, we employ a two-stage funnel annotation strategy, combining LLM-based and human annotation.
In the first stage, GPT-4o~\citep{gpt4}, performs initial annotations, followed by a second stage, where crowd-sourced human annotators conduct quality control and filtering to refine the final results.
The overall data collection pipeline is illustrated in Figure~\ref{fig:annoatation}.

\paragraph{LLM-based Annotation}
The collection pipeline is outlined as follows:
\begin{itemize}[itemsep=0.0pt,topsep=0.1pt]
  \item \textbf{Step1}: Traverse official websites recursively, collecting information on accessible sub-links and their respective pages.
  \item \textbf{Step2}: Construct queries based on the provided page information and specified role, such as focusing on the solo page or considering both pages simultaneously.
  \item \textbf{Step3}: Verify and filter for legitimate queries that deviate from natural, human-like phrasing, retaining only QA pairs with short answers containing entities.
\end{itemize}

The additional details, including step-specific prompts and case examples, are provided in Appendix~\ref{app:annotation}.
As illustrated in Figure~\ref{fig:annoatation} (b), our dataset construction includes both \texttt{multi-source} and \texttt{single-source} types, corresponding to two types of human information-seeking behaviours within web pages.
The \texttt{single-source} type simulates a user deeply exploring a single piece of information hidden within web pages, while the \texttt{multi-source} type simulates \texttt{multi-source} scenarios where users rely on multiple pages to solve a query.
Notably, the \texttt{multi-source} QA tasks can not be easily exploited by search engine shortcuts~\citep{mavi2024multihopquestionanswering}.

\paragraph{Human Annotation} 
After the synthetic queries are generated by LLM, human annotators can rewrite and calibrate the questions and answers to ensure the QA pairs are correct and consistent.

\subsection{Data Statistics} 
\label{sec:data_statistics}
Through such data construction method with LLM and human participation, we obtain 680 question-answer pairs for WebWalkerQA.
The annotated case is shown in Figure~\ref{fig:annotated}.
We will provide comprehensive statistics on WebWalkerQA, categorized by type, domain, and language.

\begin{wraptable}{r}{7.5cm}
\begin{adjustbox}{width=0.9\linewidth}
\small
\centering
\begin{tabular}{@{}c|c|c|c|c|c@{}}
\toprule
 \multicolumn{3}{c}{\texttt{Single-source} QAs} & \multicolumn{3}{c}{\texttt{Multi-source} QAs} 
\\
\toprule
\symboleeasysingle   &  \symbolmediumsingle & \symbolehardsingle & \symboleeasymuti   &  \symbolmediummuti & \symbolehardmuti \\
 Easy  & Medium & Hard  &  Easy& Medium & Hard 
\\
\midrule
80 & 140 & 120 & 80 & 140 & 120 \\
\bottomrule
\end{tabular}
\end{adjustbox}
\caption{Dataset statistics on data difficulty level.}
\label{table:comparison}
\end{wraptable}

\paragraph{Type} 
WebWalkerQA contains two types of data: \texttt{multi-source} and \texttt{single-source} QAs.
\texttt{Single-source} QAs are labeled as ${single\_source}_i$, where $i \in [2,4]$, denoting the depth of the corresponding subpage. 
Similarly, \texttt{Multi-source} QAs are labeled as ${multi\_source}_i$, where $i \in [2, 8]$, representing the sum of the depths of the two associated subpages\footnote{Taking ${multi\_source}_6$ as an example, it may refer to a query constructed from two 3rd level pages or from one page at the 2nd level and another at the 4th level.
}.
In other words, answering this query requires reading both pages simultaneously.

\paragraph{Difficulty Level}  We categorize the questions into three difficulty levels: easy, medium, and hard, based on the value of $i$. Specifically, ${single\_source}_2$, ${single\_source}_3$, and ${single\_source}_4$ correspond to the easy, medium, and hard levels, respectively.
Similarly, for multi-source questions, ${multi\_source}_{2-4}$, ${multi\_source}_{4-6}$, and ${multi\_source}_{6-8}$ correspond to the easy, medium, and hard levels, respectively.
The data statistics for the different data types are presented in Table~\ref{table:comparison}.

\paragraph{Domain} WebWalkerQA encompasses four real-world domains: conference, organization, education, and game.
These domains are selected because they provide authoritative information relevant to their respective fields, and their pages contain rich clickable content, offering substantial depth for exploration.

\begin{wrapfigure}{r}{7.5cm}
 \centering
 \includegraphics[width=0.42\textwidth]{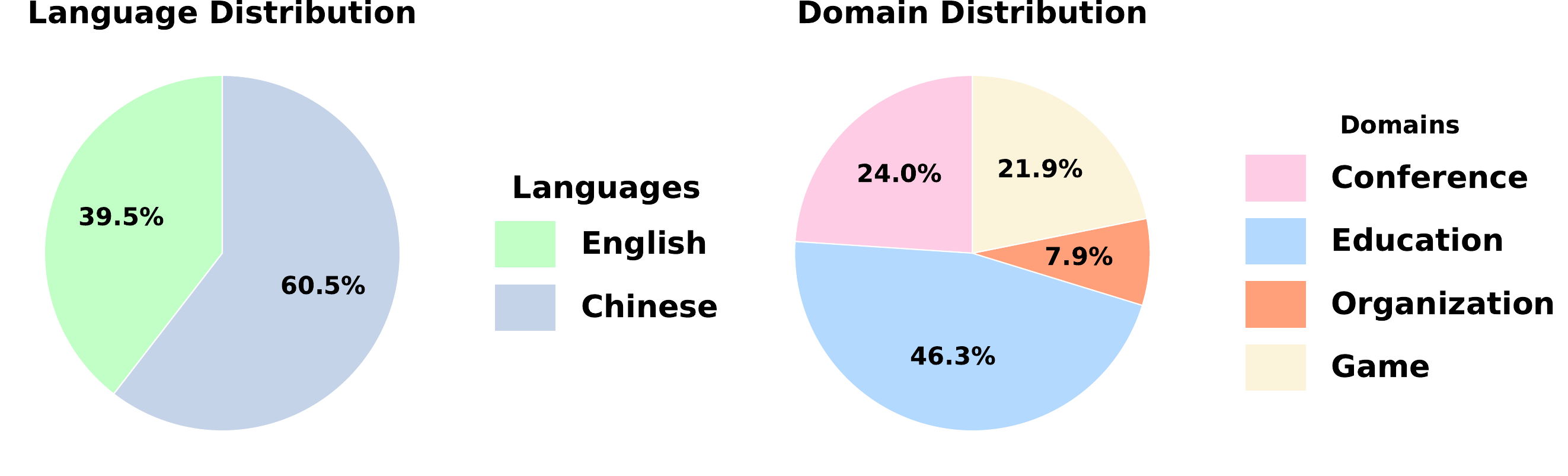}
 \caption{The language and domain distribution.  }
\label{fig:distribution}
\end{wrapfigure}
\paragraph{Language} WebWalkerQA is a bilingual dataset that includes both Chinese and English\footnote{Classification based on the language of the root webpages.}, 
reflecting the most widely used and universal languages in real-world web environments.

The statistics of WebWalkerQA on domain and language are illustrated in Figure~\ref{fig:distribution}.
The proportions of the conference, organization, education, and game domains are 24.0\%, 7.9\%, 46.3\%, and 24.0\%, respectively.
In terms of language distribution, Chinese and English account for 60.5\%, 39.5\%, respectively.
WebWalkerQA features a diverse distribution of languages and domains to ensure a comprehensive evaluation.

\subsection{Web Traversal Task and Evaluation}
\label{sec:evaluation}
Formally, given an initial website URL $U_{root}$ and a query $\mathcal{Q}$, which needs to be answered by exploring the website.
The goal of this task is to gather enough information through page traversal to ultimately answer the query $\mathcal{Q}$.
The task is to navigate the website to find the corresponding information.

WebWalkerQA can be evaluated from both \textit{performance} and \textit{efficiency} perspectives.
using question-answering \textit{accuracy} (\textit{acc.}) as the performance metric and the \textit{action count} (\textit{A.C.}) of successful agentic executions answering correctly as the efficiency metric.
Due to the varying lengths of generated text, it is challenging to perform exact match evaluation, even though we have controlled for short answers.
We use GPT-4 as the evaluator, which determines the \textit{correctness} of responses by comparing the predicted answer with the ground truth using CoT prompting strategy~\citep{wei2022chain}\footnote{ \url{https://api.python.langchain.com/en/latest/langchain/evaluation.html}, Details of the prompt for the evaluator are provided in Appendix~\ref{app:eval}}.

\section{WebWalker}
\begin{figure*}[htbp]
 \centering
 \includegraphics[width=0.87\textwidth]{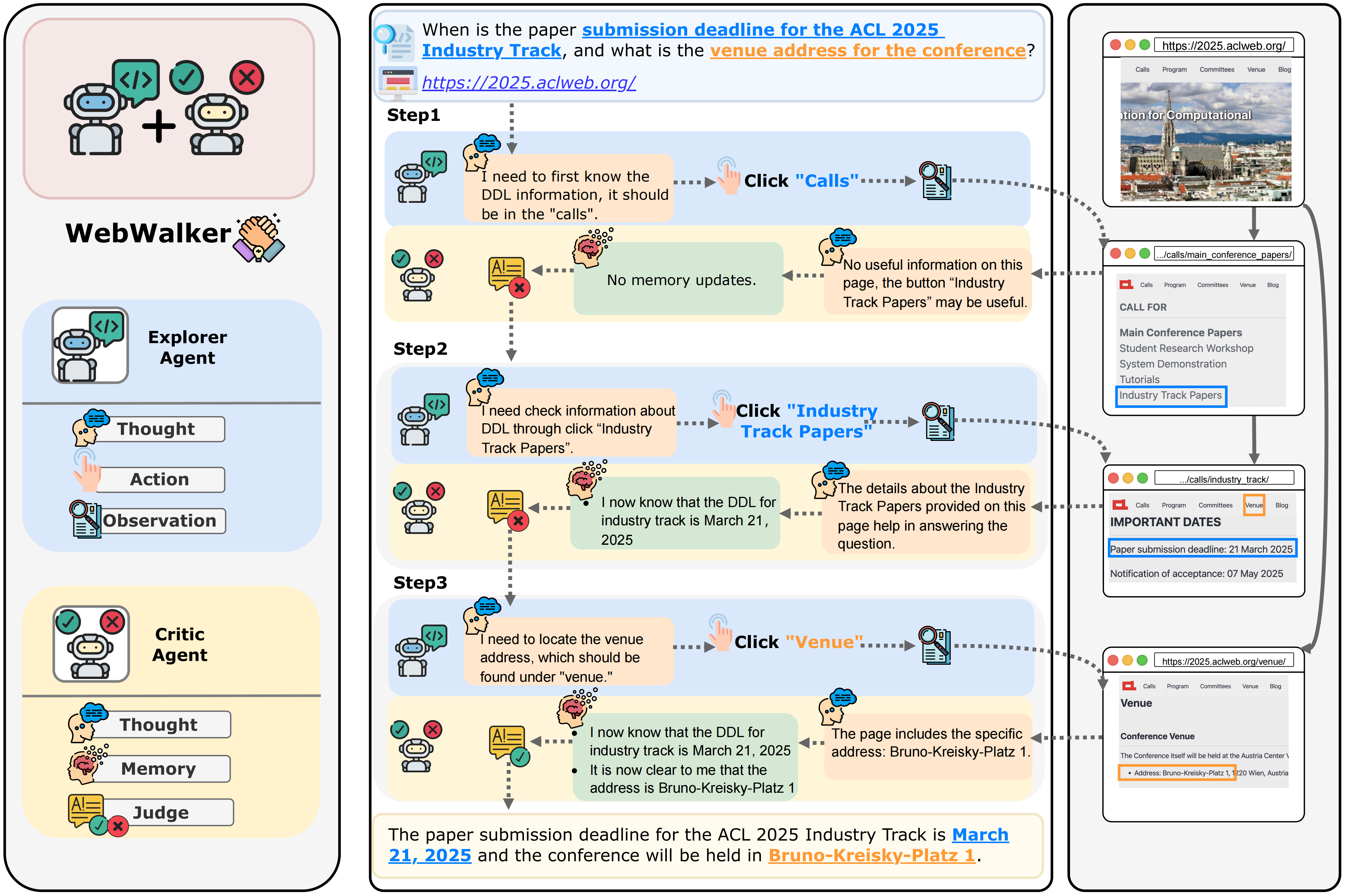}
 \caption{The overall framework of \textbf{WebWalker}.}
\label{fig:WebWalker}
\end{figure*} 
We introduce \textbf{WebWalker}, a multi-agent framework designed to interact with web environments to answer queries.
The WebWalker framework consists of two agents: an explorer agent and a critic agent.
As illustrated in Figure~\ref{fig:WebWalker}, the explorer agent traverses the web pages in Thought-Action-Observation  ($\mathcal{T}, \mathcal{A}, \mathcal{O}$) paradigms.
The critic agent updates the memory until sufficient information is accumulated to effectively address the query.
The details regarding prompts for both agents are presented in Appendix~\ref{app:prompt_agent}.


\subsection{Think then Explore}
The explorer agent explores the subpages by interacting with HTML buttons on the page.
At time step $t$, the explorer agent receives an observation $\mathcal{O}_t$ from the web environment and takes an action $\mathcal{A}_t$, following the policy $\pi(\mathcal{A}_t | \mathcal{H}_t)$.
The observation $\mathcal{O}_t = (p_t,l_t)$ consists of the information from the current page $p_t$ and a set of clickable sublinks ${l_t = \{button_i\}}_{i=1}^K$, where each $button_i$ describes HTML button information for one of the $K$ sublinks and have an associated  URL.
The action $\mathcal{A}_t$ involves selecting a URL of a subpage to explore and does \textbf{not} encompass answering the question.
Specifically, we utilize the web page's markdown content along with clickable HTML buttons (and corresponding URL)  extracted using Beautiful Soup as the observation for the current page.
The context $\mathcal{H}_t = (\mathcal{T}_1, \mathcal{A}_1, \mathcal{O}_1, \cdots, \mathcal{O}_{t-1}, \mathcal{T}_{t}, \mathcal{A}_{t}, \mathcal{O}_t)$ represents the sequence of past observations and actions leading up to the current step $t$.
The context will be updated, and this exploration process will continue until the critic agent determines to answer the query or the maximum number of steps is reached.

\subsection{Think then Critique}
Due to the policy $\pi(\mathcal{A}_t | \mathcal{H}_t)$ being implicit and the potentially large size of $\mathcal{H}_t$, motivated by pair programming~\citep{williams2000strengthening,noori2015simulation}, we incorporate a critic agent into the WebWalker framework to address these challenges.
The critic agent operates after each execution of the explorer agent.
Its input consists of the query and the explorer's current observation.
The critic initializes a memory to incrementally accumulate relevant information.
Formally, at each step, $t$, following the execution of the explorer agent, the critic agent takes the query $\mathcal{Q}$ and the explorer's current observation and action $(\mathcal{O}_t, \mathcal{A}_t)$ as input.
It then updates the memory $\mathcal{M}$, evaluates whether the gathered information is sufficiently complete to answer the query, and provides an answer once the required information is deemed sufficient.
\section{Experiment}
\subsection{Experimental Setting}
\paragraph{Baselines}
We choose widely recognized state-of-the-art agent frameworks, ReAct and Reflexion, as our baselines.
\textbf{ReAct}~\citep{react} is a general paradigm that combines reasoning and acting with LLMs by multiple thought-action-observation steps. \textbf{Reflexion}~\citep{reflexion} is a single-agent framework designed to reinforce language agents through feedback.

\paragraph{Backbones}

To thoroughly assess the web traversal capabilities of existing LLM-based agents, we select models with a context window of at least 128K to accommodate the extensive length of page information. 
Given the inherent complexity of the task, we opt for models with at least 7B parameters.
We validate a total number of nine models, including both closed-sourced and open-sourced ones:
\\
\noindent \textbf{\textit{Closed-sourced LLMs}} GPT-4o\footnote{\url{https://platform.openai.com/docs/models\#gpt-4o}} \citep{gpt4};
Qwen-Plus\footnote{\url{https://www.alibabacloud.com/help/en/model-studio/}}~\citep{qwen2.5}; \textbf{\textit{Open-sourced LLMs}} 
Qwen2.5 series models~\citep{qwen2}  specifically, Qwen2.5-\{7,14,32,72\}B-Instruct.\footnote{The LLaMA series models \citep{llama3} demonstrate limited ability to handle react-format instructions in our preliminary experiments.}

\paragraph{Implementation Details}
Considering the context limitation of models, our proposed WebWalker, along with two baselines, all operate in a zero-shot setting.
We limit the number of actions $K$ for the explorer agent to 15, meaning that the explorer agent can explore at most 15 steps.
More implementation details are presented in Appendix~\ref{app:implementation}.

\subsection{Main Results}
\begin{table*}[t]
\small
\centering
\resizebox{\columnwidth}{!}{%
\begin{tabular}{@{}l|c|cc|cc|cc|cc|cc|cc|cc@{}}
\toprule
& & \multicolumn{6}{c|}{\texttt{Single-source} QA} & \multicolumn{6}{c|}{\texttt{Multi-source} QA} & \multicolumn{2}{c}{\multirow{3}*{\textbf{Overall}}} \\
\cmidrule{3-14}
\multirow{2}*{\textbf{Backbones}} & \multirow{2}*{\textbf{Method}} & \multicolumn{2}{c}{\symboleeasysingle}   &  \multicolumn{2}{c}{\symbolmediumsingle} & \multicolumn{2}{c|}{\symbolehardsingle} & \multicolumn{2}{c}{\symboleeasymuti}   &  \multicolumn{2}{c}{\symbolmediummuti} & \multicolumn{2}{c|}{\symbolehardmuti} \\
&& \multicolumn{2}{c}{Easy}  &  \multicolumn{2}{c}{Medium} & \multicolumn{2}{c|}{Hard} & \multicolumn{2}{c}{Easy}    &  \multicolumn{2}{c}{Medium} & \multicolumn{2}{c|}{Hard} & \\
\cmidrule{3-16}
& &\textit{acc.} & \textit{A.C.}  &\textit{acc.} & \textit{A.C.} &\textit{acc.} & \textit{A.C.}  &\textit{acc.} & \textit{A.C.}  &\textit{acc.} & \textit{A.C.}  &\textit{acc.} & \textit{A.C.} &\textit{acc.} & \textit{A.C.}   \\
\midrule
\multicolumn{16}{c}{\cellcolor{uclablue} \textbf{\textit{Closed-Sourced LLMs}}} \\
\midrule
\multirow{3}{*}{GPT-4o} 
& ReAct & 53.75 & 2.53 & 45.00 & 3.34 & 30.00 & 5.61 & 32.50 & 2.34 & 31.43 & 3.97 & 15.00 & 6.77 & 33.82 & 3.83\\
 & Reflexion & 56.25 & 2.91 & 51.43 & 3.88 & 30.83 & 5.75 & 35.00 & 3.67 & 27.14 & 4.13 & 16.67 & 7.05 &35.29 & 4.27\\
 & WebWalker & 55.00 & 2.97 & 50.00 & 3.43 & 30.00 &
 6.02 & 47.50 & 4.00 & 34.29 & 3.85 & 15.83 & 6.57 &\textbf{37.50} & 4.67\\
\arrayrulecolor{black}\midrule
\multirow{3}{*}{Qwen-Plus}
& ReAct & 48.75 & 1.67 & 48.57 &2.69 & 28.33 & 4.00& 35.00 & 2.60 & 27.86 & 3.11 & 14.17 & 6.55 & 33.08 & 3.03\\
 & Reflexion & 53.75 & 3.66 & 40.00 & 3.79 & 24.17 & 5.88 & 47.50 & 3.28 & 30.00 & 4.07 & 15.00 & 7.11 & 33.23 & 4.32\\
 & WebWalker & 55.00 & 3.72 & 47.14 & 3.19 & 30.00 & 6.13 & 35.00 & 3.89 & 27.14 & 4.39 & 15.00 & 7.38 & \textbf{33.82} & 4.36\\
\arrayrulecolor{black}\midrule
\multicolumn{16}{c}{\cellcolor{uclagold} \textbf{\textit{Open-Sourced LLMs}}} \\
\midrule
\multirow{3}{*}{\makecell[l]{Qwen-2.5\\-7B}} 
& ReAct & 37.50 & 3.36 & 18.5 7& 4.88  & 9.17 & 5.45 & 17.50 & 3.42 & 11.43 & 3.62 & 5.83 & 4.57 & 16.02 &2.99\\
 & Reflexion & 37.50 & 4.03 & 25.00 & 3.48 & 11.67 & 4.57 & 30.00 & 2.66 & 15.71 & 5.45 & 4.17 & 7.8 & 19.11 & 4.07\\
 & WebWalker & 41.25 & 3.39 & 24.71 & 3.86 & 12.50 & 5.93 & 18.75 & 3.00 & 20.71 & 3.34 & 5.83 & 7.28 & \textbf{19.85} & 3.94\\
\arrayrulecolor{black!20}\midrule
\multirow{3}{*}{\makecell[l]{Qwen-2.5\\-14B}}
& ReAct & 36.25 & 1.86 & 32.14 & 2.75 & 15.00 & 3.61 & 27.50 & 2.31 & 22.86 & 3.00 & 5.00 & 5.00 & 22.35 & 2.76\\
 & Reflexion & 46.25 & 2.21 & 34.29& 2.83 & 15.00& 4.44 & 36.25 & 2.51 & 22.86 & 3.34 & 5.83 & 5.42 & 25.14 & 3.01\\
 & WebWalker & 41.25 & 2.42 & 41.43 & 3.24 & 23.33 & 4.42  & 30.00 & 3.95 & 22.86 & 3.56 & 10.00 & 6.16 & \textbf{27.50} & 3.60\\
\arrayrulecolor{black!20}\midrule
\multirow{3}{*}{\makecell[l]{Qwen-2.5\\-32B}}
& ReAct & 47.50 & 2.21 & 35.71 & 3.20 & 16.67 & 3.55 & 36.25 & 2.68 & 18.57 & 3.00 & 8.33 & 3.70 & 25.44 & 2.93\\
 & Reflexion & 42.50 & 2.52 & 32.86 & 2.65 & 16.67 & 3.90 & 31.25 & 2.84 & 23.57 & 3.12 & 5.83 & 5.00 & 23.26 & 3.00\\
 & WebWalker & 41.25 & 2.69 & 34.29 & 4.14 & 22.50 & 5.14 & 27.50 & 3.13 & 25.00 & 3.51 & 10.00 & 6.08 & \textbf{26.02} & 3.90\\
\arrayrulecolor{black!20}\midrule
\multirow{3}{*}{\makecell[l]{Qwen-2.5\\-72B}} 
& ReAct & 47.50 & 1.68 & 38.57 & 2.79 & 20.00 & 4.04 & 45.00 & 2.25 & 32.14 & 3.13 & 10.00 & 5.41 & 30.73 & 2.86\\
 & Reflexion & 57.50 & 3.04 & 44.29 & 3.88 & 28.33 & 5.82 & 36.25 & 3.62 & 25.00 & 3.60 & 12.50 & 6.26 & 32.50 & 4.09\\
 & WebWalker & 58.75 &  2.70 & 48.57 & 3.07 & 25.83 & 5.77 & 35.00 & 3.57 & 29.29 & 4.87 & 15.00 & 7.38 & \textbf{33.26} & 4.32\\
\arrayrulecolor{black}\bottomrule
\end{tabular}%
}
\caption{Main results of three methods across closed-sourced and open-sourced LLMs as the backbone. 
\textit{Acc.} and \textit{A.C.} refer to accuracy and action count, respectively.
}
\label{table:main_result}
\end{table*}

\label{sec:main_result}
The main results across six LLMs are presented in Table~\ref{table:main_result}. 
The closed-source models outperform the open-source models in both performance and efficiency.
For open-source models, performance and efficiency improves as the model size increases.
Our proposed WebWalker framework outperforms Reflexion, which in turn outperforms React.
We only counted the action count (\textit{A.C.}) from correct executions, and as the model size increases, the \textit{A.C.} grows, indicating that larger LLMs have enhanced long-range information-seeking ability.
Even the best-performing WebWalker using GPT-4o as its backbone does not surpass 40\%, highlighting the challenge posed by WebWalkerQA.
It can be observed that as the depth increases or the number of sources required increases, the difficulty of acquiring the information needed to resolve the query becomes greater, resulting in a decline in accuracy performance.

\begin{figure}[t]
    \centering
    \subfigure[]{
        \includegraphics[width=0.31\textwidth]{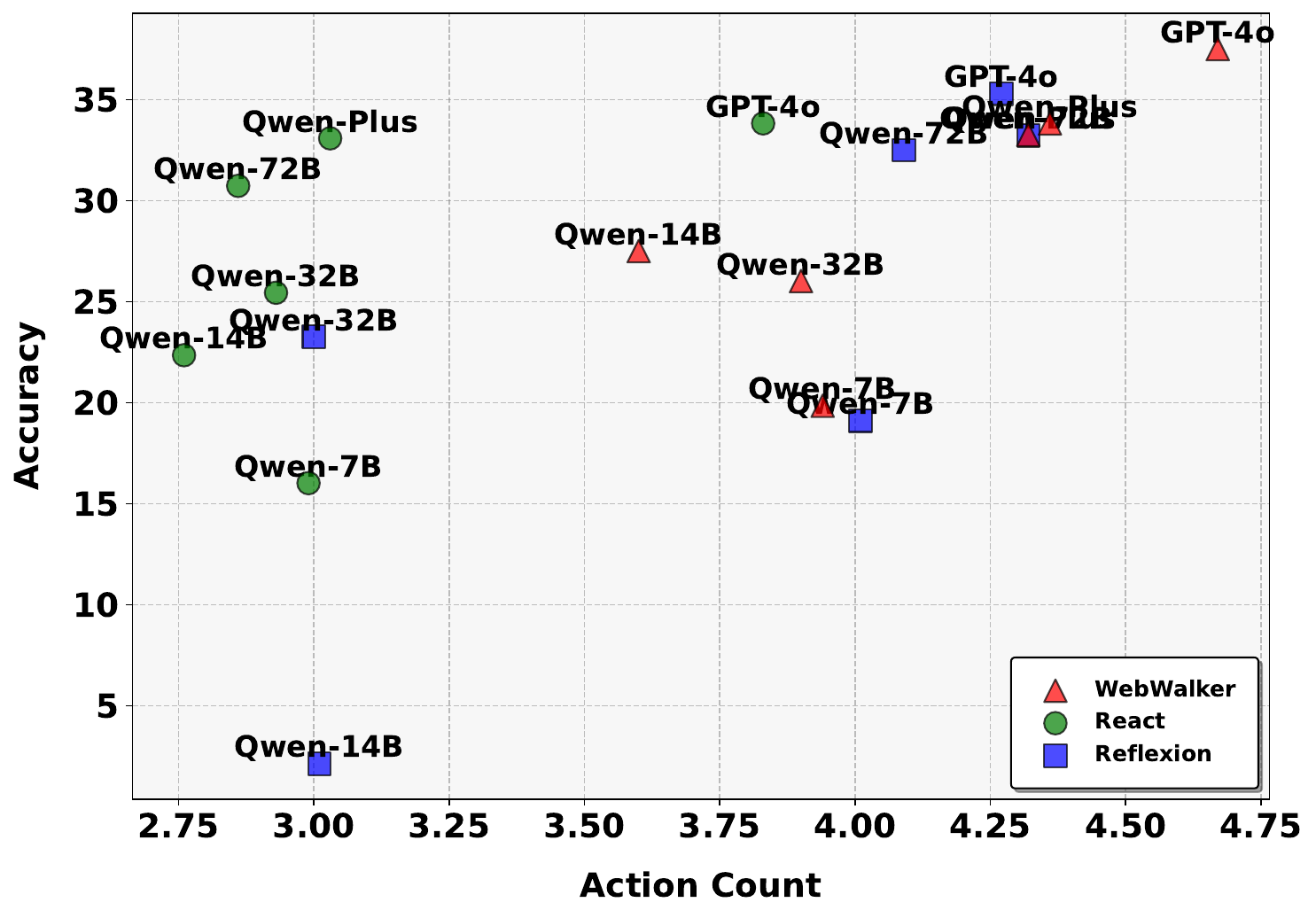}
    }
    \subfigure[]{
        \includegraphics[width=0.62\textwidth]{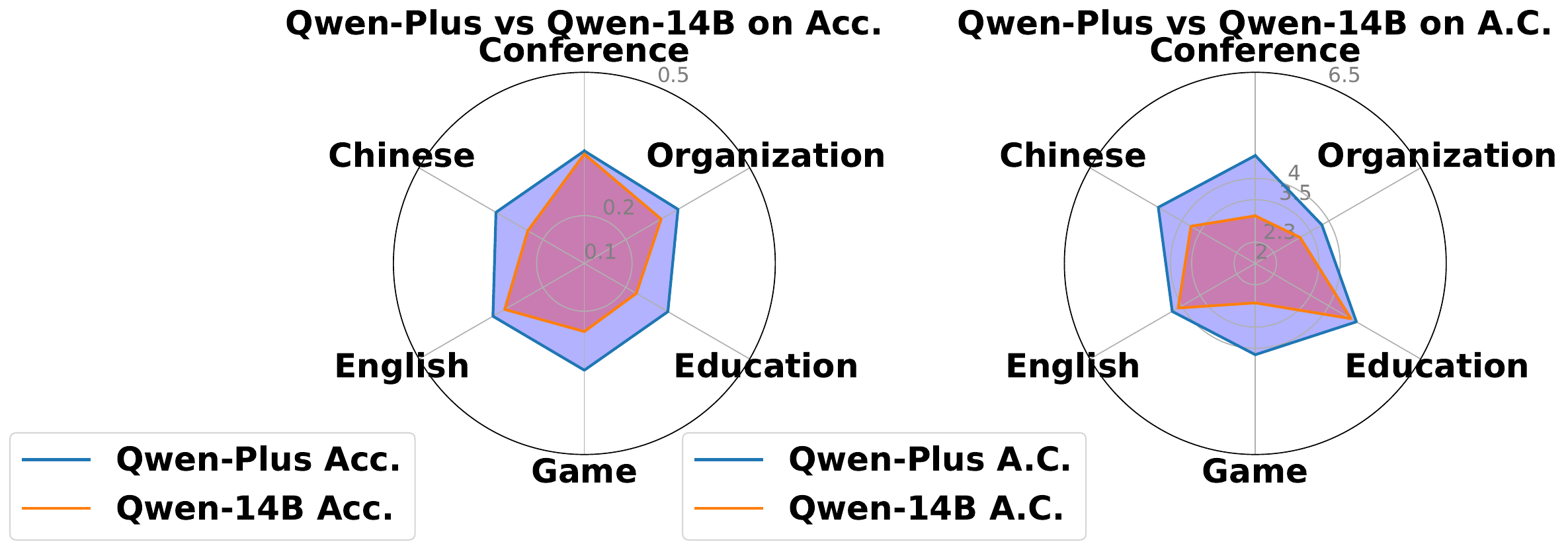}
    }
    \caption{(a) \textcolor{red!70!transparent}{$\blacktriangle$} represents WebWalker using various models as backbones, 
    \textcolor{blue!70!transparent}{$\blacksquare$} represents Reflextion with different backbone models, and 
    \textcolor{green!70!transparent}{\large $\bullet$} denotes ReAct employing various backbone models.
    (b) Performance across domains and languages of WebWalker building upon Qwen-14B and Qwen-Plus.}
    \label{table:detailed_result}
\end{figure}

The performance distribution of accuracy and action count for different methods across various models is shown in Figure~\ref{fig:result_dis}.
The further towards the top-right corner, the more effective and prolonged the web traversal becomes.
We observe that increasing the model size or introducing reflection on the process of each action can address certain problems requiring multi-step solutions, thereby enabling long-distance task-solving capabilities in web traversal tasks.

\subsection{Results across Domains and Languages}

WebWalkerQA is a bilingual dataset encompassing both Chinese and English and spans multiple domains, including games, conferences, education, and organizations.
The performance across different domains and languages is shown in Figure~\ref{fig:detailed_result}.
In the domain of \textbf{conference}, the framework demonstrates relatively superior performance, likely due to the more explicit and directive nature of the button information, which facilitates more straightforward inferences.
The framework performs similarly in both Chinese and English, as the models we employed are both pre-trained and supervised-fine-tuned in a bilingual setting.

\subsection{Error Assessment}
\begin{wrapfigure}{r}{7.5cm}
 \centering
 \includegraphics[width=0.48\textwidth]{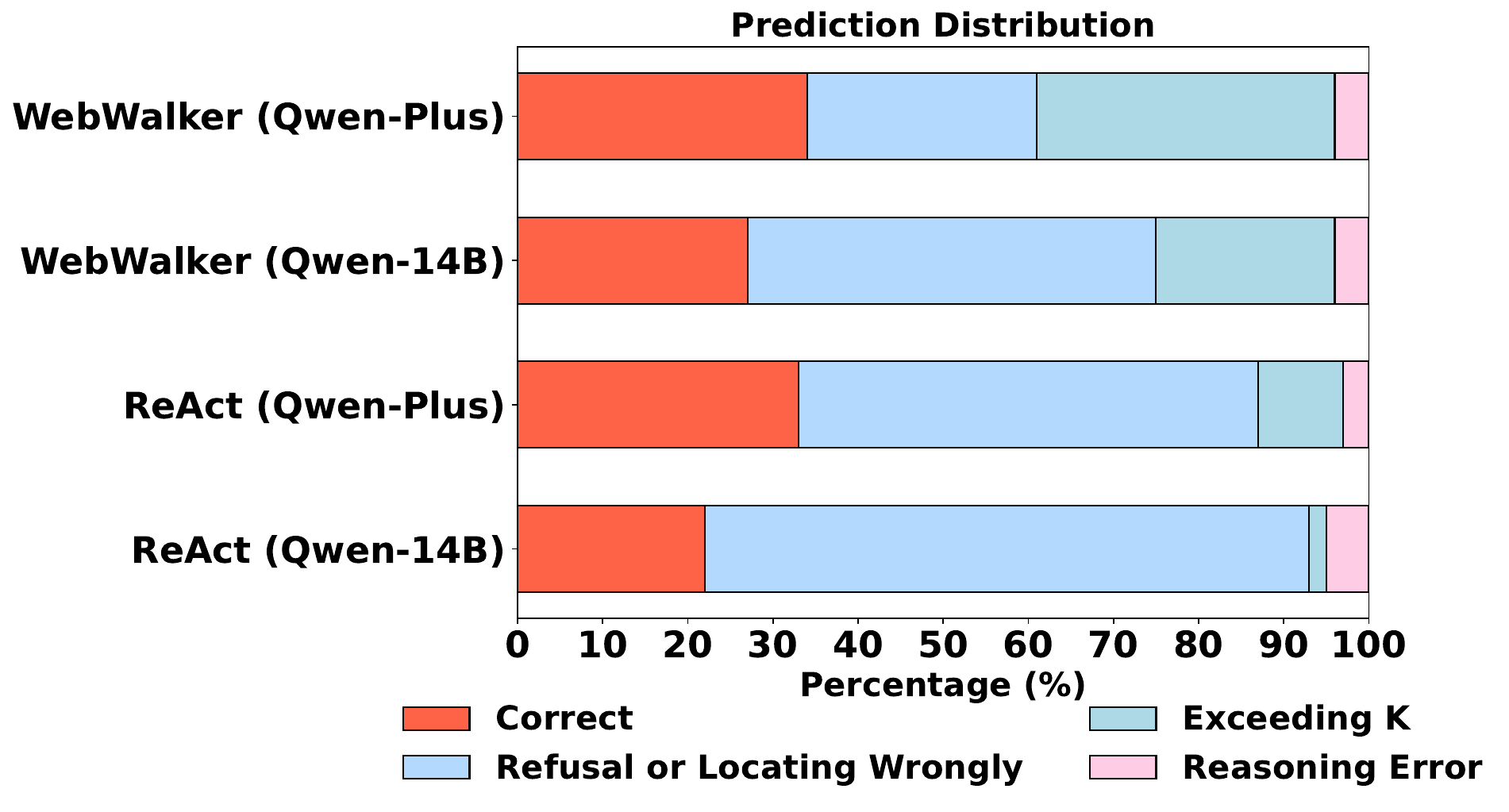}
 \caption{Predication distribution of WebWalker and React method building on Qwen-14B and Qwen-Plus.}
 
\label{fig:error}
\vspace{-4mm}
\end{wrapfigure}

For incorrect execution, errors can also be categorized into three types: refusal to answer or locating wrongly, reasoning error, and exceeding the maximum number of steps $K$.
The prediction distribution is shown in Figure~\ref{fig:error}.
The model with a relatively small number of parameters using the ReAct framework lacks the capacity to explore the depth of information, making judgments within just a few iterations of taking action, regardless of whether relevant information has been found.
It tends to ``\textit{\underline{give up}}'' and exhibits characteristics of \textit{\underline{impatience}}.
Introducing memory to manage the long context, along with an increase in model parameters, provides evidence that this phenomenon stems from the interference of long contexts having noisy information and the inherent capabilities of the model itself, consistent with the analysis drawn in \S\ref{sec:main_result}.
Some errors are categorized as reasoning errors, where the golden page has been found in the visited pages but is still incorrectly marked.
This underscores the challenge of reasoning on page information in certain cases.\footnote{The corresponding case is presented in Appendix~\ref{sec:reasonerror}.}
\section{Discussion}
\begin{table*}[t]
\scriptsize
\centering
\begin{tabular}{@{}l|c|c|c|c|c|c|c@{}}
\toprule
\multirow{3}*{\textbf{Systems}}& \multicolumn{3}{c|}{\texttt{Single-source} QA} & \multicolumn{3}{c|}{\texttt{Multi-source} QA} & \multicolumn{1}{c}{\multirow{3}*{\textbf{Overall}}} \\
& \multicolumn{1}{c}{\symboleeasysingle}   &  \multicolumn{1}{c}{\symbolmediumsingle} & \multicolumn{1}{c|}{\symbolehardsingle} & \multicolumn{1}{c}{\symboleeasymuti}   &  \multicolumn{1}{c}{\symbolmediummuti} & \multicolumn{1}{c|}{\symbolehardmuti} \\
  & \multicolumn{1}{c}{Easy}  &  \multicolumn{1}{c}{Medium} & \multicolumn{1}{c|}{Hard} & \multicolumn{1}{c}{Easy}    &  \multicolumn{1}{c}{Medium} & \multicolumn{1}{c|}{Hard} & \\
\midrule
\multicolumn{8}{c}{\cellcolor{babygreen} \textbf{\textit{ Close Book (No Retrieval)}}} \\
\midrule
Gemini-1.5-Pro & 12.50 & 7.86 & 8.33 & 11.25 & 6.43 & 5.00 & 8.08 \\
o1-preview & 16.25 & 10.00 & 9.17 & 7.50 & 10.71 & 6.67 & 9.85\\
\midrule
\multicolumn{8}{c}{\cellcolor{babyred} \textbf{\textit{Commerical Systems}}} \\
\midrule
Doubao& 45.00 & 15.00 & 18.33 & 13.75 & 8.57 & 10.00 & 16.76 \\
Gemini-Search & 40.00 & 32.14 & 29.17 & 30.00 & 23.57 & 17.50 & 27.94 \\
ERNIE-4.0-8K & 52.50 & 30.00 & 28.33 & 21.25 & 18.57 & 30.00 & 28.97\\
Kimi & 77.50 & 41.43 & 40.83 & 26.25 & 26.43& 22.50 & 37.35\\
Tongyi & 41.25 & 45.00 & 41.67 & 40.00 & 41.43 & 34.17 & 40.73\\
\midrule
\multicolumn{8}{c}{\cellcolor{aliceblue} \textbf{\textit{Open-Sourced Systems}}} \\
\midrule
Naive RAG & 37.50 & 25.71 & 24.17 & 20.00 & 14.29 & 12.50 & 20.73 \\
MindSearch & 15.00 & 11.43 & 10.83 & 8.75 & 12.14 & 10.00 & 11.32\\
\midrule
\cellcolor{mycell}{\textbf{Avg.}} & \cellcolor{mycell}{37.50} & \cellcolor{mycell}{24.29} & \cellcolor{mycell}{23.42} &\cellcolor{mycell}{19.86} & \cellcolor{mycell}{18.02} & \cellcolor{mycell}{16.48} & \cellcolor{mycell}{-}\\
\bottomrule
\end{tabular}
\caption{Accuracy results on Commercial and Open-sourced Searched-enhanced RAG systems.}
\label{table:rag_result}
\vspace{-4mm}
\end{table*}

\subsection{RAG Performance on WebWalkerQA}
\label{sec:rag}
We evaluate the performance of RAG systems in tackling WebWalkerQA's challenges, specifically, whether they can retrieve \underline{deep} information, presented in Table~\ref{table:rag_result}.

We first evaluate the performance under \textbf{\textit{Close Book}} settings
using the state-of-the-art model OpenAI o1~\citep{o1} and Gemini-1.5-Pro without retrieval.
We then access the performance of several commercial and open-sourced RAG systems\footnote{The commercial RAG systems are accessed through business-oriented API. The details of RAG systems are provided in Appendix~\ref{app:rag}.}.
Without performing the search, even the strongest models exhibit very poor performance.
WebWalkerQA is built on official websites with dynamically updated information, while pre-trained models rely on static knowledge limited by a cutoff date and lack dynamic updates\footnote{The case study is shown in Appendix~\ref{sec:time_cutoff}.}.
Both commercial and open-sourced RAG systems exhibit relatively poor performance on WebWalkerQA, with the best result coming from Tongyi, which only reaches 40\%.
Commercial RAG systems are typically modular, consisting of various components such as rewrite, router, reranker, and others.
Some systems, like ERNIE, may have stronger search capabilities for Chinese, resulting in higher values.
For open-sourced RAG systems, \texttt{Multi-source} queries have lower accuracy than \texttt{Single-source} queries, which validates the challenge posed by WebWalkerQA, as search engines are \textbf{unable} to retrieve all relevant information in one or several single horizontal search attempts.
Furthermore, as the difficulty increases, $\eg$ the depth of information growing deeper, the performance tends to deteriorate. 
Overall, search engines still face challenges when retrieving content that is buried deeper.
\begin{findings}
\textbf{\underline{Findings (i)}}: \textit{RAG systems struggle with key challenges that require effective web traversal.}
\end{findings}

\subsection{WebWalker Combined with RAG System}
\label{sec:combine}


\begin{figure}[t]
    \centering
    \subfigure[\label{fig:combine}]{
        \includegraphics[width=0.48\textwidth]{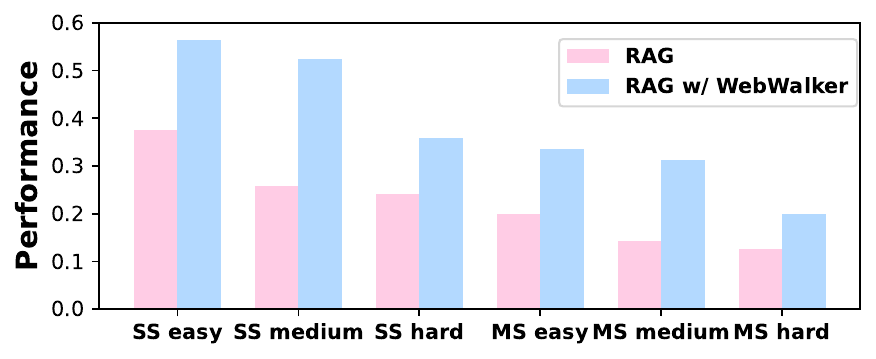}
    }\hfill
    \subfigure[\label{fig:scale}]{
        \includegraphics[width=0.41\textwidth]{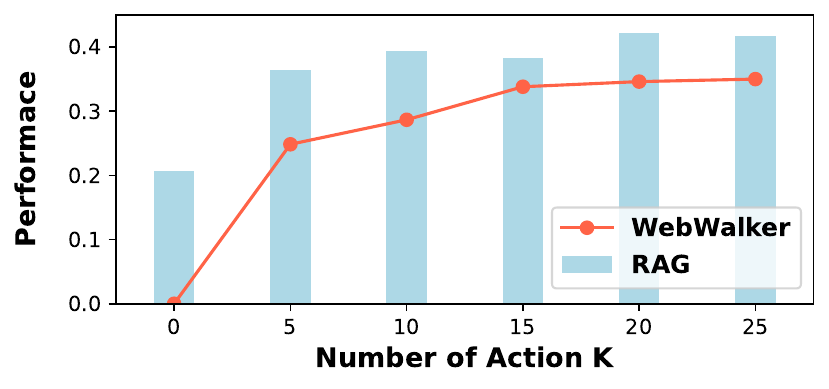}
    }
    \caption{(a) Performance under standard RAG and RAG combined with WebWalker configurations. \textbf{SS} and \textbf{MS} denote \texttt{single-source} and \texttt{multi-source} QAs.
    (b) Overall performance on WebWalker and RAG combined WebWalker at varying values of $K$, using Qwen-Plus as backbones.}
    \vspace{-6mm}
\end{figure}

The standard RAG system can be viewed as a \textit{\underline{horizontal search}} for relevant documents in response to a query, while WebWalker can be considered as a vertical exploration approach.
WebWalker can seamlessly integrate into standard RAG systems to acquire deep information and enhance problem-solving capabilities.
We integrate WebWalker building upon Qwen-2.5-Plus into the naive RAG system, and the detailed results are shown in Figure~\ref{fig:combine}.
The core contribution of WebWalker is providing useful information for question answering; specifically, the memory $\mathcal{M}$ of the critic agent is append to the relevant documents to aid in generation.
It is observed that, after the integration, performance has improved across all difficulty levels, especially in the \texttt{multi-source} category.
\begin{findings}
\textbf{\underline{Findings (ii)}}: \textit{WebWalker can be a module in agentic RAG system, enabling \textit{vertical exploration}
}.
\end{findings}

\subsection{Scaling Up on Action Count $K$}
\label{sec:scaleup}
Previous work~\citep{yue2024inference} explored the inference scaling laws for the RAG system by examining the impact of increasing retrieved
documents.
We scale up the amount of $K \in \{5, 10, 15, 20, 25\}$ to study the impact of scaling during the inference phase when tracing source information.
Figure~\ref{fig:scale} shows the results of scaling up, where larger values of $K$ lead to better performance, validating the feasibility of vertical scaling within a certain range.

\begin{findings}
\textbf{\underline{Findings (iii)}}: \textit{Scaling the process of digging through links could represent a potential direction for vertical exploration in RAG systems.}
\end{findings}
\section{Conclusion}
We introduce WebWalkerQA, a benchmark for evaluating LLMs' web traversal abilities in complex, multi-step information-seeking tasks. 
We also proposed WebWalker, a multi-agent framework that mimics human-like web navigation, combining exploration and critique. 
Experiments show that WebWalkerQA effectively challenges RAG systems, and combining RAG with WebWalker improves web navigation performance. 
Our work highlights the importance of deep, vertical exploration in web-based tasks, paving the way for more scalable and reliable LLM-based information retrieval integrated with RAG.

\clearpage
\bibliography{biblio}

\begin{thebibliography}{55}
\providecommand{\natexlab}[1]{#1}
\providecommand{\url}[1]{\texttt{#1}}
\expandafter\ifx\csname urlstyle\endcsname\relax
  \providecommand{\doi}[1]{doi: #1}\else
  \providecommand{\doi}{doi: \begingroup \urlstyle{rm}\Url}\fi

\bibitem[Abuelsaad et~al.(2024)Abuelsaad, Akkil, Dey, Jagmohan, Vempaty, and Kokku]{abuelsaad2024agent}
Tamer Abuelsaad, Deepak Akkil, Prasenjit Dey, Ashish Jagmohan, Aditya Vempaty, and Ravi Kokku.
\newblock Agent-e: From autonomous web navigation to foundational design principles in agentic systems.
\newblock In \emph{NeurIPS 2024 Workshop on Open-World Agents}, 2024.

\bibitem[Chen et~al.(2024{\natexlab{a}})Chen, Liu, Wang, Liu, Zhang, Chen, and Zhao]{mindsearch}
Zehui Chen, Kuikun Liu, Qiuchen Wang, Jiangning Liu, Wenwei Zhang, Kai Chen, and Feng Zhao.
\newblock Mindsearch: Mimicking human minds elicits deep ai searcher.
\newblock \emph{arXiv preprint arXiv:2407.20183}, 2024{\natexlab{a}}.

\bibitem[Chen et~al.(2024{\natexlab{b}})Chen, Liu, Wang, Zhang, Liu, Lin, Chen, and Zhao]{chen-etal-2024-agent}
Zehui Chen, Kuikun Liu, Qiuchen Wang, Wenwei Zhang, Jiangning Liu, Dahua Lin, Kai Chen, and Feng Zhao.
\newblock Agent-{FLAN}: Designing data and methods of effective agent tuning for large language models.
\newblock In Lun-Wei Ku, Andre Martins, and Vivek Srikumar (eds.), \emph{Findings of the Association for Computational Linguistics: ACL 2024}, pp.\  9354--9366, Bangkok, Thailand, August 2024{\natexlab{b}}. Association for Computational Linguistics.
\newblock \doi{10.18653/v1/2024.findings-acl.557}.
\newblock URL \url{https://aclanthology.org/2024.findings-acl.557}.

\bibitem[Cheng et~al.(2024)Cheng, Sun, Chu, Xu, Li, Zhang, and Wu]{cheng2024seeclick}
Kanzhi Cheng, Qiushi Sun, Yougang Chu, Fangzhi Xu, Yantao Li, Jianbing Zhang, and Zhiyong Wu.
\newblock Seeclick: Harnessing gui grounding for advanced visual gui agents.
\newblock \emph{arXiv preprint arXiv:2401.10935}, 2024.

\bibitem[Deng et~al.(2023)Deng, Gu, Zheng, Chen, Stevens, Wang, Sun, and Su]{deng2023mindweb}
Xiang Deng, Yu~Gu, Boyuan Zheng, Shijie Chen, Samuel Stevens, Boshi Wang, Huan Sun, and Yu~Su.
\newblock Mind2web: Towards a generalist agent for the web.
\newblock In \emph{Thirty-seventh Conference on Neural Information Processing Systems Datasets and Benchmarks Track}, 2023.
\newblock URL \url{https://openreview.net/forum?id=kiYqbO3wqw}.

\bibitem[Deng et~al.(2024)Deng, Zhang, Zhang, Yuan, Ng, and Chua]{mt_mind2web_agent}
Yang Deng, Xuan Zhang, Wenxuan Zhang, Yifei Yuan, See-Kiong Ng, and Tat-Seng Chua.
\newblock On the multi-turn instruction following for conversational web agents.
\newblock In Lun-Wei Ku, Andre Martins, and Vivek Srikumar (eds.), \emph{Proceedings of the 62nd Annual Meeting of the Association for Computational Linguistics (Volume 1: Long Papers)}, pp.\  8795--8812, Bangkok, Thailand, August 2024. Association for Computational Linguistics.
\newblock \doi{10.18653/v1/2024.acl-long.477}.
\newblock URL \url{https://aclanthology.org/2024.acl-long.477}.

\bibitem[Dubey et~al.(2024)Dubey, Jauhri, Pandey, Kadian, Al{-}Dahle, Letman, Mathur, Schelten, Yang, Fan, Goyal, Hartshorn, Yang, Mitra, Sravankumar, Korenev, Hinsvark, Rao, Zhang, Rodriguez, Gregerson, and et~al.]{llama3}
Abhimanyu Dubey, Abhinav Jauhri, Abhinav Pandey, Abhishek Kadian, Ahmad Al{-}Dahle, Aiesha Letman, Akhil Mathur, Alan Schelten, Amy Yang, Angela Fan, Anirudh Goyal, Anthony Hartshorn, Aobo Yang, Archi Mitra, Archie Sravankumar, Artem Korenev, Arthur Hinsvark, Arun Rao, Aston Zhang, Aur{\'{e}}lien Rodriguez, Austen Gregerson, and et~al.
\newblock The llama 3 herd of models.
\newblock \emph{CoRR}, abs/2407.21783, 2024.
\newblock \doi{10.48550/ARXIV.2407.21783}.
\newblock URL \url{https://doi.org/10.48550/arXiv.2407.21783}.

\bibitem[Furuta et~al.(2024)Furuta, Lee, Nachum, Matsuo, Faust, Gu, and Gur]{furuta2024multimodal}
Hiroki Furuta, Kuang-Huei Lee, Ofir Nachum, Yutaka Matsuo, Aleksandra Faust, Shixiang~Shane Gu, and Izzeddin Gur.
\newblock Multimodal web navigation with instruction-finetuned foundation models.
\newblock In \emph{The Twelfth International Conference on Learning Representations}, 2024.
\newblock URL \url{https://openreview.net/forum?id=efFmBWioSc}.

\bibitem[He et~al.(2024{\natexlab{a}})He, Yao, Ma, Yu, Dai, Zhang, Lan, and Yu]{he-etal-2024-webvoyager}
Hongliang He, Wenlin Yao, Kaixin Ma, Wenhao Yu, Yong Dai, Hongming Zhang, Zhenzhong Lan, and Dong Yu.
\newblock {W}eb{V}oyager: Building an end-to-end web agent with large multimodal models.
\newblock In Lun-Wei Ku, Andre Martins, and Vivek Srikumar (eds.), \emph{Proceedings of the 62nd Annual Meeting of the Association for Computational Linguistics (Volume 1: Long Papers)}, pp.\  6864--6890, Bangkok, Thailand, August 2024{\natexlab{a}}. Association for Computational Linguistics.
\newblock \doi{10.18653/v1/2024.acl-long.371}.
\newblock URL \url{https://aclanthology.org/2024.acl-long.371}.

\bibitem[He et~al.(2024{\natexlab{b}})He, Yao, Ma, Yu, Zhang, Fang, Lan, and Yu]{he2024openwebvoyager}
Hongliang He, Wenlin Yao, Kaixin Ma, Wenhao Yu, Hongming Zhang, Tianqing Fang, Zhenzhong Lan, and Dong Yu.
\newblock Openwebvoyager: Building multimodal web agents via iterative real-world exploration, feedback and optimization.
\newblock \emph{arXiv preprint arXiv:2410.19609}, 2024{\natexlab{b}}.

\bibitem[Hong et~al.(2024)Hong, Wang, Lv, Xu, Yu, Ji, Wang, Wang, Dong, Ding, et~al.]{hong2024cogagent}
Wenyi Hong, Weihan Wang, Qingsong Lv, Jiazheng Xu, Wenmeng Yu, Junhui Ji, Yan Wang, Zihan Wang, Yuxiao Dong, Ming Ding, et~al.
\newblock Cogagent: A visual language model for gui agents.
\newblock In \emph{Proceedings of the IEEE/CVF Conference on Computer Vision and Pattern Recognition}, pp.\  14281--14290, 2024.

\bibitem[Humphreys et~al.(2022)Humphreys, Raposo, Pohlen, Thornton, Chhaparia, Muldal, Abramson, Georgiev, Santoro, and Lillicrap]{humphreys2022data}
Peter~C Humphreys, David Raposo, Tobias Pohlen, Gregory Thornton, Rachita Chhaparia, Alistair Muldal, Josh Abramson, Petko Georgiev, Adam Santoro, and Timothy Lillicrap.
\newblock A data-driven approach for learning to control computers.
\newblock In \emph{International Conference on Machine Learning}, pp.\  9466--9482. PMLR, 2022.

\bibitem[Iong et~al.(2024)Iong, Liu, Chen, Lai, Yao, Shen, Yu, Dong, and Tang]{iong2024openwebagent}
Iat~Long Iong, Xiao Liu, Yuxuan Chen, Hanyu Lai, Shuntian Yao, Pengbo Shen, Hao Yu, Yuxiao Dong, and Jie Tang.
\newblock Openwebagent: An open toolkit to enable web agents on large language models.
\newblock In \emph{Proceedings of the 62nd Annual Meeting of the Association for Computational Linguistics (Volume 3: System Demonstrations)}, pp.\  72--81, 2024.

\bibitem[Kim et~al.(2024)Kim, Baldi, and McAleer]{kim2024language}
Geunwoo Kim, Pierre Baldi, and Stephen McAleer.
\newblock Language models can solve computer tasks.
\newblock \emph{Advances in Neural Information Processing Systems}, 36, 2024.

\bibitem[Koh et~al.(2024{\natexlab{a}})Koh, Lo, Jang, Duvvur, Lim, Huang, Neubig, Zhou, Salakhutdinov, and Fried]{koh2024visualwebarena}
Jing~Yu Koh, Robert Lo, Lawrence Jang, Vikram Duvvur, Ming~Chong Lim, Po-Yu Huang, Graham Neubig, Shuyan Zhou, Ruslan Salakhutdinov, and Daniel Fried.
\newblock Visualwebarena: Evaluating multimodal agents on realistic visual web tasks.
\newblock In \emph{ICLR 2024 Workshop on Large Language Model (LLM) Agents}, 2024{\natexlab{a}}.

\bibitem[Koh et~al.(2024{\natexlab{b}})Koh, McAleer, Fried, and Salakhutdinov]{treesearchagent}
Jing~Yu Koh, Stephen McAleer, Daniel Fried, and Ruslan Salakhutdinov.
\newblock Tree search for language model agents.
\newblock \emph{arXiv preprint arXiv:2407.01476}, 2024{\natexlab{b}}.

\bibitem[Lai et~al.(2024)Lai, Liu, Iong, Yao, Chen, Shen, Yu, Zhang, Zhang, Dong, et~al.]{lai2024autowebglm}
Hanyu Lai, Xiao Liu, Iat~Long Iong, Shuntian Yao, Yuxuan Chen, Pengbo Shen, Hao Yu, Hanchen Zhang, Xiaohan Zhang, Yuxiao Dong, et~al.
\newblock Autowebglm: A large language model-based web navigating agent.
\newblock In \emph{Proceedings of the 30th ACM SIGKDD Conference on Knowledge Discovery and Data Mining}, pp.\  5295--5306, 2024.

\bibitem[Lewis et~al.(2020)Lewis, Perez, Piktus, Petroni, Karpukhin, Goyal, K{\"u}ttler, Lewis, Yih, Rockt{\"a}schel, et~al.]{rag}
Patrick Lewis, Ethan Perez, Aleksandra Piktus, Fabio Petroni, Vladimir Karpukhin, Naman Goyal, Heinrich K{\"u}ttler, Mike Lewis, Wen-tau Yih, Tim Rockt{\"a}schel, et~al.
\newblock Retrieval-augmented generation for knowledge-intensive nlp tasks.
\newblock \emph{Advances in Neural Information Processing Systems}, 33:\penalty0 9459--9474, 2020.

\bibitem[Liu et~al.(2018)Liu, Guu, Pasupat, Shi, and Liang]{liu2018reinforcement}
Evan~Zheran Liu, Kelvin Guu, Panupong Pasupat, Tianlin Shi, and Percy Liang.
\newblock Reinforcement learning on web interfaces using workflow-guided exploration.
\newblock \emph{arXiv preprint arXiv:1802.08802}, 2018.

\bibitem[Liu et~al.(2023)Liu, Lai, Yu, Xu, Zeng, Du, Zhang, Dong, and Tang]{liu2023webglm}
Xiao Liu, Hanyu Lai, Hao Yu, Yifan Xu, Aohan Zeng, Zhengxiao Du, Peng Zhang, Yuxiao Dong, and Jie Tang.
\newblock Webglm: Towards an efficient web-enhanced question answering system with human preferences.
\newblock In \emph{Proceedings of the 29th ACM SIGKDD Conference on Knowledge Discovery and Data Mining}, pp.\  4549--4560, 2023.

\bibitem[Liu et~al.(2024)Liu, Yu, Zhang, Xu, Lei, Lai, Gu, Ding, Men, Yang, Zhang, Deng, Zeng, Du, Zhang, Shen, Zhang, Su, Sun, Huang, Dong, and Tang]{liuagentbench}
Xiao Liu, Hao Yu, Hanchen Zhang, Yifan Xu, Xuanyu Lei, Hanyu Lai, Yu~Gu, Hangliang Ding, Kaiwen Men, Kejuan Yang, Shudan Zhang, Xiang Deng, Aohan Zeng, Zhengxiao Du, Chenhui Zhang, Sheng Shen, Tianjun Zhang, Yu~Su, Huan Sun, Minlie Huang, Yuxiao Dong, and Jie Tang.
\newblock Agentbench: Evaluating {LLM}s as agents.
\newblock In \emph{The Twelfth International Conference on Learning Representations}, 2024.
\newblock URL \url{https://openreview.net/forum?id=zAdUB0aCTQ}.

\bibitem[Mavi et~al.(2024)Mavi, Jangra, and Jatowt]{mavi2024multihopquestionanswering}
Vaibhav Mavi, Anubhav Jangra, and Adam Jatowt.
\newblock Multi-hop question answering, 2024.
\newblock URL \url{https://arxiv.org/abs/2204.09140}.

\bibitem[Mialon et~al.(2024)Mialon, Fourrier, Wolf, LeCun, and Scialom]{mialon2024gaia}
Gr{\'e}goire Mialon, Cl{\'e}mentine Fourrier, Thomas Wolf, Yann LeCun, and Thomas Scialom.
\newblock {GAIA}: a benchmark for general {AI} assistants.
\newblock In \emph{The Twelfth International Conference on Learning Representations}, 2024.
\newblock URL \url{https://openreview.net/forum?id=fibxvahvs3}.

\bibitem[Nakano et~al.(2021)Nakano, Hilton, Balaji, Wu, Ouyang, Kim, Hesse, Jain, Kosaraju, Saunders, et~al.]{nakano2021webgpt}
Reiichiro Nakano, Jacob Hilton, Suchir Balaji, Jeff Wu, Long Ouyang, Christina Kim, Christopher Hesse, Shantanu Jain, Vineet Kosaraju, William Saunders, et~al.
\newblock Webgpt: Browser-assisted question-answering with human feedback.
\newblock \emph{arXiv preprint arXiv:2112.09332}, 2021.

\bibitem[Nguyen et~al.(2024)Nguyen, Chen, Wang, Wu, Park, Hu, Lyu, Wu, Aponte, Xia, et~al.]{nguyen2024gui}
Dang Nguyen, Jian Chen, Yu~Wang, Gang Wu, Namyong Park, Zhengmian Hu, Hanjia Lyu, Junda Wu, Ryan Aponte, Yu~Xia, et~al.
\newblock Gui agents: A survey.
\newblock \emph{arXiv preprint arXiv:2412.13501}, 2024.

\bibitem[Noori \& Kazemifard(2015)Noori and Kazemifard]{noori2015simulation}
Fariba Noori and Mohammad Kazemifard.
\newblock Simulation of pair programming using multi-agent and mbti personality model.
\newblock In \emph{2015 Sixth International Conference of Cognitive Science (ICCS)}, pp.\  29--36. IEEE, 2015.

\bibitem[OpenAI(2022{\natexlab{a}})]{chatgpt}
OpenAI.
\newblock Introducing {ChatGPT}, 2022{\natexlab{a}}.
\newblock URL \url{https://openai.com/blog/chatgpt}.

\bibitem[OpenAI(2022{\natexlab{b}})]{gpt4}
OpenAI.
\newblock Gpt-4 system card, 2022{\natexlab{b}}.
\newblock URL \url{https://cdn.openai.com/papers/gpt-4-system-card.pdf}.

\bibitem[OpenAI(2024)]{o1}
OpenAI.
\newblock Introducing openai o1, 2024.
\newblock URL \url{https://openai.com/o1/}.

\bibitem[Ouyang et~al.(2022)Ouyang, Wu, Jiang, Almeida, Wainwright, Mishkin, Zhang, Agarwal, Slama, Ray, et~al.]{ouyang2022training}
Long Ouyang, Jeffrey Wu, Xu~Jiang, Diogo Almeida, Carroll Wainwright, Pamela Mishkin, Chong Zhang, Sandhini Agarwal, Katarina Slama, Alex Ray, et~al.
\newblock Training language models to follow instructions with human feedback.
\newblock \emph{Advances in neural information processing systems}, 35:\penalty0 27730--27744, 2022.

\bibitem[Qiao et~al.(2024)Qiao, Zhang, Fang, Luo, Zhou, Jiang, Lv, and Chen]{qiao-etal-2024-autoact}
Shuofei Qiao, Ningyu Zhang, Runnan Fang, Yujie Luo, Wangchunshu Zhou, Yuchen Jiang, Chengfei Lv, and Huajun Chen.
\newblock {A}uto{A}ct: Automatic agent learning from scratch for {QA} via self-planning.
\newblock In Lun-Wei Ku, Andre Martins, and Vivek Srikumar (eds.), \emph{Proceedings of the 62nd Annual Meeting of the Association for Computational Linguistics (Volume 1: Long Papers)}, pp.\  3003--3021, Bangkok, Thailand, August 2024. Association for Computational Linguistics.
\newblock \doi{10.18653/v1/2024.acl-long.165}.
\newblock URL \url{https://aclanthology.org/2024.acl-long.165}.

\bibitem[Reddy et~al.(2024)Reddy, Mukherjee, Kim, Wang, Hakkani-Tur, and Ji]{reddy2024infogent}
Revanth~Gangi Reddy, Sagnik Mukherjee, Jeonghwan Kim, Zhenhailong Wang, Dilek Hakkani-Tur, and Heng Ji.
\newblock Infogent: An agent-based framework for web information aggregation.
\newblock \emph{arXiv preprint arXiv:2410.19054}, 2024.

\bibitem[Shinn et~al.(2024)Shinn, Cassano, Gopinath, Narasimhan, and Yao]{reflexion}
Noah Shinn, Federico Cassano, Ashwin Gopinath, Karthik Narasimhan, and Shunyu Yao.
\newblock Reflexion: Language agents with verbal reinforcement learning.
\newblock \emph{Advances in Neural Information Processing Systems}, 36, 2024.

\bibitem[Song et~al.(2024)Song, Xu, Zhou, and Neubig]{song2024beyond}
Yueqi Song, Frank Xu, Shuyan Zhou, and Graham Neubig.
\newblock Beyond browsing: Api-based web agents.
\newblock \emph{arXiv preprint arXiv:2410.16464}, 2024.

\bibitem[Tan et~al.(2024)Tan, Dou, Wang, Wang, Chen, and Wen]{tan2024htmlrag}
Jiejun Tan, Zhicheng Dou, Wen Wang, Mang Wang, Weipeng Chen, and Ji-Rong Wen.
\newblock Htmlrag: Html is better than plain text for modeling retrieved knowledge in rag systems.
\newblock \emph{arXiv preprint arXiv:2411.02959}, 2024.

\bibitem[Team(2024)]{qwen2.5}
Qwen Team.
\newblock Qwen2.5: A party of foundation models, September 2024.
\newblock URL \url{https://qwenlm.github.io/blog/qwen2.5/}.

\bibitem[Wei et~al.(2022)Wei, Wang, Schuurmans, Bosma, Xia, Chi, Le, Zhou, et~al.]{wei2022chain}
Jason Wei, Xuezhi Wang, Dale Schuurmans, Maarten Bosma, Fei Xia, Ed~Chi, Quoc~V Le, Denny Zhou, et~al.
\newblock Chain-of-thought prompting elicits reasoning in large language models.
\newblock \emph{Advances in neural information processing systems}, 35:\penalty0 24824--24837, 2022.

\bibitem[Williams et~al.(2000)Williams, Kessler, Cunningham, and Jeffries]{williams2000strengthening}
Laurie Williams, Robert~R Kessler, Ward Cunningham, and Ron Jeffries.
\newblock Strengthening the case for pair programming.
\newblock \emph{IEEE software}, 17\penalty0 (4):\penalty0 19--25, 2000.

\bibitem[Xu et~al.(2024)Xu, Kordi, Nayak, Asija, Wang, Sanders, Byerly, Zhang, Durme, and Khashabi]{xu2024turkingbenchchallengebenchmarkweb}
Kevin Xu, Yeganeh Kordi, Tanay Nayak, Ado Asija, Yizhong Wang, Kate Sanders, Adam Byerly, Jingyu Zhang, Benjamin~Van Durme, and Daniel Khashabi.
\newblock Tur[k]ingbench: A challenge benchmark for web agents, 2024.
\newblock URL \url{https://arxiv.org/abs/2403.11905}.

\bibitem[Xu et~al.(2021)Xu, Masling, Du, Campagna, Heck, Landay, and Lam]{xu-etal-2021-grounding}
Nancy Xu, Sam Masling, Michael Du, Giovanni Campagna, Larry Heck, James Landay, and Monica Lam.
\newblock Grounding open-domain instructions to automate web support tasks.
\newblock In Kristina Toutanova, Anna Rumshisky, Luke Zettlemoyer, Dilek Hakkani-Tur, Iz~Beltagy, Steven Bethard, Ryan Cotterell, Tanmoy Chakraborty, and Yichao Zhou (eds.), \emph{Proceedings of the 2021 Conference of the North American Chapter of the Association for Computational Linguistics: Human Language Technologies}, pp.\  1022--1032, Online, June 2021. Association for Computational Linguistics.
\newblock \doi{10.18653/v1/2021.naacl-main.80}.
\newblock URL \url{https://aclanthology.org/2021.naacl-main.80}.

\bibitem[Yang et~al.(2024)Yang, Yang, Hui, Zheng, Yu, Zhou, Li, Li, Liu, Huang, Dong, Wei, Lin, Tang, Wang, Yang, Tu, Zhang, Ma, Yang, Xu, Zhou, and et~al.]{qwen2}
An~Yang, Baosong Yang, Binyuan Hui, Bo~Zheng, Bowen Yu, Chang Zhou, Chengpeng Li, Chengyuan Li, Dayiheng Liu, Fei Huang, Guanting Dong, Haoran Wei, Huan Lin, Jialong Tang, Jialin Wang, Jian Yang, Jianhong Tu, Jianwei Zhang, Jianxin Ma, Jianxin Yang, Jin Xu, Jingren Zhou, and et~al.
\newblock Qwen2 technical report.
\newblock \emph{CoRR}, abs/2407.10671, 2024.
\newblock \doi{10.48550/ARXIV.2407.10671}.
\newblock URL \url{https://doi.org/10.48550/arXiv.2407.10671}.

\bibitem[Yao et~al.(2022)Yao, Chen, Yang, and Narasimhan]{yao2022webshop}
Shunyu Yao, Howard Chen, John Yang, and Karthik Narasimhan.
\newblock Webshop: Towards scalable real-world web interaction with grounded language agents.
\newblock \emph{Advances in Neural Information Processing Systems}, 35:\penalty0 20744--20757, 2022.

\bibitem[Yao et~al.(2023)Yao, Zhao, Yu, Du, Shafran, Narasimhan, and Cao]{react}
Shunyu Yao, Jeffrey Zhao, Dian Yu, Nan Du, Izhak Shafran, Karthik~R Narasimhan, and Yuan Cao.
\newblock React: Synergizing reasoning and acting in language models.
\newblock In \emph{The Eleventh International Conference on Learning Representations}, 2023.
\newblock URL \url{https://openreview.net/forum?id=WE_vluYUL-X}.

\bibitem[Yoran et~al.(2024)Yoran, Amouyal, Malaviya, Bogin, Press, and Berant]{assistantbench}
Ori Yoran, Samuel~Joseph Amouyal, Chaitanya Malaviya, Ben Bogin, Ofir Press, and Jonathan Berant.
\newblock {A}ssistant{B}ench: Can web agents solve realistic and time-consuming tasks?
\newblock In Yaser Al-Onaizan, Mohit Bansal, and Yun-Nung Chen (eds.), \emph{Proceedings of the 2024 Conference on Empirical Methods in Natural Language Processing}, pp.\  8938--8968, Miami, Florida, USA, November 2024. Association for Computational Linguistics.
\newblock URL \url{https://aclanthology.org/2024.emnlp-main.505}.

\bibitem[Yue et~al.(2024)Yue, Zhuang, Bai, Hui, Jagerman, Zeng, Qin, Wang, Wang, and Bendersky]{yue2024inference}
Zhenrui Yue, Honglei Zhuang, Aijun Bai, Kai Hui, Rolf Jagerman, Hansi Zeng, Zhen Qin, Dong Wang, Xuanhui Wang, and Michael Bendersky.
\newblock Inference scaling for long-context retrieval augmented generation.
\newblock \emph{arXiv preprint arXiv:2410.04343}, 2024.

\bibitem[Zeng et~al.(2024)Zeng, Liu, Lu, Wang, Liu, Dong, and Tang]{zeng-etal-2024-agenttuning}
Aohan Zeng, Mingdao Liu, Rui Lu, Bowen Wang, Xiao Liu, Yuxiao Dong, and Jie Tang.
\newblock {A}gent{T}uning: Enabling generalized agent abilities for {LLM}s.
\newblock In Lun-Wei Ku, Andre Martins, and Vivek Srikumar (eds.), \emph{Findings of the Association for Computational Linguistics: ACL 2024}, pp.\  3053--3077, Bangkok, Thailand, August 2024. Association for Computational Linguistics.
\newblock \doi{10.18653/v1/2024.findings-acl.181}.
\newblock URL \url{https://aclanthology.org/2024.findings-acl.181}.

\bibitem[Zhang et~al.(2024{\natexlab{a}})Zhang, He, Qian, Li, Li, Qin, Kang, Ma, Lin, Rajmohan, et~al.]{zhang2024large}
Chaoyun Zhang, Shilin He, Jiaxu Qian, Bowen Li, Liqun Li, Si~Qin, Yu~Kang, Minghua Ma, Qingwei Lin, Saravan Rajmohan, et~al.
\newblock Large language model-brained gui agents: A survey.
\newblock \emph{arXiv preprint arXiv:2411.18279}, 2024{\natexlab{a}}.

\bibitem[Zhang et~al.(2024{\natexlab{b}})Zhang, Tang, Wu, Wang, Shen, Hou, Tan, Li, Zhuang, and Lu]{zhang2024agent}
Wenqi Zhang, Ke~Tang, Hai Wu, Mengna Wang, Yongliang Shen, Guiyang Hou, Zeqi Tan, Peng Li, Yueting Zhuang, and Weiming Lu.
\newblock Agent-pro: Learning to evolve via policy-level reflection and optimization.
\newblock \emph{arXiv preprint arXiv:2402.17574}, 2024{\natexlab{b}}.

\bibitem[Zhang et~al.(2024{\natexlab{c}})Zhang, Tian, Chen, and Liu]{zhang2024mmina}
Ziniu Zhang, Shulin Tian, Liangyu Chen, and Ziwei Liu.
\newblock Mmina: Benchmarking multihop multimodal internet agents, 2024{\natexlab{c}}.
\newblock URL \url{https://arxiv.org/abs/2404.09992}.

\bibitem[Zheng et~al.(2024{\natexlab{a}})Zheng, Gou, Kil, Sun, and Su]{seeact}
Boyuan Zheng, Boyu Gou, Jihyung Kil, Huan Sun, and Yu~Su.
\newblock Gpt-4v (ision) is a generalist web agent, if grounded.
\newblock In \emph{Forty-first International Conference on Machine Learning}, 2024{\natexlab{a}}.

\bibitem[Zheng et~al.(2024{\natexlab{b}})Zheng, Gou, Salisbury, Du, Sun, and Su]{zheng-etal-2024-webolympus}
Boyuan Zheng, Boyu Gou, Scott Salisbury, Zheng Du, Huan Sun, and Yu~Su.
\newblock {W}eb{O}lympus: An open platform for web agents on live websites.
\newblock In Delia~Irazu Hernandez~Farias, Tom Hope, and Manling Li (eds.), \emph{Proceedings of the 2024 Conference on Empirical Methods in Natural Language Processing: System Demonstrations}, pp.\  187--197, Miami, Florida, USA, November 2024{\natexlab{b}}. Association for Computational Linguistics.
\newblock URL \url{https://aclanthology.org/2024.emnlp-demo.20}.

\bibitem[Zhou et~al.(2024{\natexlab{a}})Zhou, Xu, Zhu, Zhou, Lo, Sridhar, Cheng, Ou, Bisk, Fried, Alon, and Neubig]{webarena}
Shuyan Zhou, Frank~F. Xu, Hao Zhu, Xuhui Zhou, Robert Lo, Abishek Sridhar, Xianyi Cheng, Tianyue Ou, Yonatan Bisk, Daniel Fried, Uri Alon, and Graham Neubig.
\newblock Webarena: A realistic web environment for building autonomous agents.
\newblock In \emph{The Twelfth International Conference on Learning Representations}, 2024{\natexlab{a}}.
\newblock URL \url{https://openreview.net/forum?id=oKn9c6ytLx}.

\bibitem[Zhou et~al.(2023)Zhou, Jiang, Li, Wu, Wang, Qiu, Zhang, Chen, Wu, Wang, et~al.]{zhou2023agents}
Wangchunshu Zhou, Yuchen~Eleanor Jiang, Long Li, Jialong Wu, Tiannan Wang, Shi Qiu, Jintian Zhang, Jing Chen, Ruipu Wu, Shuai Wang, et~al.
\newblock Agents: An open-source framework for autonomous language agents.
\newblock \emph{arXiv preprint arXiv:2309.07870}, 2023.

\bibitem[Zhou et~al.(2024{\natexlab{b}})Zhou, Ou, Ding, Li, Wu, Wang, Chen, Wang, Xu, Zhang, et~al.]{zhou2024symbolic}
Wangchunshu Zhou, Yixin Ou, Shengwei Ding, Long Li, Jialong Wu, Tiannan Wang, Jiamin Chen, Shuai Wang, Xiaohua Xu, Ningyu Zhang, et~al.
\newblock Symbolic learning enables self-evolving agents.
\newblock \emph{arXiv preprint arXiv:2406.18532}, 2024{\natexlab{b}}.

\bibitem[Zhu et~al.(2024)Zhu, Qiao, Ou, Deng, Zhang, Lyu, Shen, Liang, Gu, and Chen]{zhu2024knowagent}
Yuqi Zhu, Shuofei Qiao, Yixin Ou, Shumin Deng, Ningyu Zhang, Shiwei Lyu, Yue Shen, Lei Liang, Jinjie Gu, and Huajun Chen.
\newblock Knowagent: Knowledge-augmented planning for llm-based agents, 2024.
\newblock URL \url{https://arxiv.org/abs/2403.03101}.

\end{thebibliography}
\bibliographystyle{colm2024_conference}

\clearpage
\appendix
\section{Limitations and Discussion}
We discuss the following limitations:\\
\noindent \textbf{Dataset Size}: 
Due to the complexity of queries in the web-agent domain, similar to benchmarks such as AssistantBench~\citep{assistantbench} (214) and MMIna ~\citep{zhang2024mmina} (1,050), GAIA~\citep{mialon2024gaia} (466), our proposed WebWalkerQA currently comprises 680 high-quality QA pairs.
Additionally, we possess a collection of approximately 14k silver QA pairs, which, although not yet carefully human-verified, can serve as supplementary \textbf{training data} to enhance agent performance, leaving room for further exploration.
\\
\noindent \textbf{Multimodal Environment}: 
In this work, we only utilize HTML-DOM to parse clickable buttons.
In fact, visual modalities, such as screenshots, can also assist and provide a more intuitive approach~\citep{nguyen2024gui,zhang2024large,he2024openwebvoyager}. 
We leave this for future work. 
\\
\noindent \textbf{Agent Tuning}: 
WebWalker is driven by prompting without additional training.
We can use agent tuning to help LLMs learn web traversal. This involves fine-tuning models with golden trajectories, enabling them to take effective actions for completing information-seeking tasks~\citep{zeng-etal-2024-agenttuning,chen-etal-2024-agent, zhang2024agent,qiao-etal-2024-autoact,zhu2024knowagent}.
\\
\noindent \textbf{Better Integration with RAG Systems}: 
In \S\ref{sec:combine}, the root url is provided for the WebWalker to execute.
To better integrate with the RAG system, one approach could be to first rewrite the query within the RAG system to refine the search, directing it to the query's official websites likely to contain relevant information. 
The WebWalker can then be used to extract useful information.
Both the knowledge retrieved from the RAG system and the information mined by the WebWalker can be combined as augmented retrieval knowledge for generation, leading to a better result.

WebWalker can function independently as a \textbf{web information retrieval assistant} for a given webpage or \textbf{seamlessly integrate with RAG systems} to expand their scope. 
Under the agentic RAG paradigm, the \textit{click} \symbolclick action proves to be highly effective.
\section{Implementation Details}
\label{app:implementation}

In this study, we utilize \texttt{Qwen-Agent}\footnote{\url{https://github.com/QwenLM/Qwen-Agent}} as the foundational codebase for building and developing the baselines proposed WebWalker. 
The details of LLM hyperparameters for generation are as follows: $top_p=0.8$.
We sincerely thank the contributors and maintainers of \texttt{ai4crawl}\footnote{\url{https://github.com/unclecode/crawl4ai}} for their open-source tool, which helped us get web pages in a Markdown-like format.
We will release the code of WebWalker in \github GitHub.

\section{Details for RAG Systems}
\label{app:rag}
We select five mainstream commercial systems and two open-source systems for evaluation.
\subsection{Commercial Systems}
\label{app:rag_commercial}
Doubao\footnote{ \url{https://www.volcengine.com/docs/82379/1302004}}, ERNIE-4.0-8K\footnote{\url{https://cloud.baidu.com/doc/WENXINWORKSHOP/s/clntwmv7t}},
Tongyi, Kimi, and Gemini-Search are all accessed through their business-oriented API interfaces to ensure reproducibility.
The detailed configuration of each API can be found in our codebase.

\subsection{Open-sourced Systems}
\label{app:rag_open_source}
(a) \textbf{Mindsearch}~\citep{mindsearch} is to mimic the human
minds in web information seeking and integration, which can be instantiated by a multi-agent framework consisting of a WebPlanner and WebSearcher. 
(b) \textbf{Naive RAG built from scratch} We use Google to query the relevant terms and concatenate the information from the Top-10 returned links with the query to provide instructions for the Qwen-Plus to generate a response.

\section{Annotated Case}
An annotated case is shown in Figure~\ref{fig:annotated}.
The WebWalkerQA dataset will be available at \hfdataset HuggingFace Datasets.
\begin{figure}
\begin{tcolorbox}[title={\textbf{\small Annotated Data Format}}, colback=whitesmoke, colframe=royalblue(web), boxrule=2pt, arc=0mm]
{\small
\begin{lstlisting}[style=mystyle]
## JSON Format
The keys in the JSON include: 
Question, Answer, Root_Url, and Info. The Info field contains 
more detailed information, including Hop, Domain, Language, 
Difficulty_Level, Source Website, and Golden_Path.
```
{
    "Question": "When is the paper submission deadline for the ACL 2025 Industry Track, and what is the venue address for the conference?",
    "Answer": "The paper submission deadline for the ACL 2025 Industry Track is March 21, 2025. The conference will be held in Brune-Kreisky-Platz 1.",
    "Root_Url": "https://2025.aclweb.org/",
    "Info":{
        "Hop": "multi-source",
        "Domain": "Conference",
        "Language": "English",
        "Difficulty_Level": "Medium",
        "Source_Website": ["https://2025.aclweb.org/calls/industry_track/","https://2025.aclweb.org/venue/"],
        "Golden_Path": ["root->call>student_research_workshop", "root->venue"]
    }
}
```
\end{lstlisting}
}
\end{tcolorbox}
\caption{A JSON-format case in WebWalkerQA.}
\label{fig:annotated}
\end{figure}

\section{Details on Annotation}
\label{app:annotation}

\subsection{Sources of Root Page}
The root page is initially identified through a Google search using keywords such as ``\textit{conference official website}'' or ``\textit{game official website}'', followed by manual filtering.
For the education domain, we choose the official websites of various university computer science departments, closely reflecting real-world scenarios.
The distribution of the domain is shown in Figure~\ref{fig:distribution}.

\subsection{Details on Prompts for Annotation}
\label{app:prompt_annotation}
The prompts for GPT-4o-based initial annotation are presented below.
\begin{tcolorbox}[
enhanced jigsaw,
breakable,
pad at break*=1mm,
colback=white!95!gray,
colframe=gray!50!black,
title={Prompts for Multi-source Data Annotation}]
\small
\textbf{Question Generate}
\begin{lstlisting}[breaklines=true, xleftmargin=0pt, breakindent=0pt, columns=fullflexible]
You are a professional web content analyst. Based on the provided material, construct a query statement:
Sublink 1 URL; Sublink 1 INFO 
Sublink 2 URL; Sublink 2 INFO 
...
Sublink n URL; Sublink n INFO 
### Requirements:
1. **Core Goal of the Query**: Create a multi-step standalone query where the user needs to integrate information from at least two sublinks to find the final answer. The answer should be a single, clear, concise, and precise entity.
2. **Relevance of Sublinks**: The selected sublinks must have an intrinsic connection, and the answer should be derived by combining information from these two sublinks.
3. **Logical and Complex**: The constructed query should be as complex and specific as possible, challenging, and can leverage time, sequence, or commonly mentioned topics to construct a naturally coherent reasoning process. Avoid questions about browsing history, browsing paths, etc., which have no practical value.
4. **Accuracy of the Answer**: Ensure the answer is accurate, concise, and closely connected to the logical chain constructed in the query.

Please return in JSON format, structured as follows:
{
    "sublink_reason": "Describe why these specific sublinks were chosen and how they are interrelated.",
    "sublinks": ["Selected sublink URL", "Selected sublink URL"],
    "reason": "Explain the reason for designing this query and how it encourages the user to engage in multi-step reasoning.",
    "query": "Your query statement",
    "answer": "The answer to the query"
}
(*@\textcolor{blue}{
Sublink 1 URL; Sublink 1 INFO  \\
Sublink 2 URL; Sublink 2 INFO \\
... \\
Sublink n URL; Sublink n INFO
}@*)
\end{lstlisting}
\textbf{Question-Answer Verify}
\begin{lstlisting}[breaklines=true, xleftmargin=0pt, breakindent=0pt, columns=fullflexible]
You will act as a strict judge. You need to evaluate whether the given query can be accurately answered only by combining the information from two documents (doc1 and doc2) and the provided answer. Additionally, check if the answer is concise (as an entity or a judgment) and correct.

If the answer is incorrect, can be answered using only one document, or is not concise enough, you should return false.
If any document (doc1 or doc2) does not contain the necessary key information for the answer and only provides context for the query, you should return false.
If any document merely provides query background information unrelated to the answer and does not require combining information from both documents, you should return false.
If the answer is a long answer and not of an entity type, you should return false.
If the query is unnatural, doesn't appear as a complete query, or has a harsh tone, you should return false.
Each question should require combining information from both documents, meaning the answer results from multi-hop reasoning or multi-step reasoning, and it is concise for you to return true.
You are very strict, and any case failing to meet the above criteria should result in a false. Please return your result in JSON format as follows:
{
    "reason": "Consider each of the conditions above in sequence to assess whether the query and answer meet the criteria. If they do meet the criteria, list the helpful parts from each doc for answering the question.",
    "decision": "true/false"
}
(*@\textcolor{blue}{\{Doc1 INFO\}; \{Doc2 INFO\}
}@*)
\end{lstlisting}
\end{tcolorbox}

\begin{tcolorbox}[
enhanced jigsaw,
breakable,
pad at break*=1mm,
colback=white!95!gray,
colframe=gray!50!black,
title={Prompts for Single-source Data Annotation}]
\small
\textbf{Question Generate}
\begin{lstlisting}[breaklines=true, xleftmargin=0pt, breakindent=0pt, columns=fullflexible]

\end{lstlisting}
\textbf{Question-Answer Verify}
\begin{lstlisting}[breaklines=true, xleftmargin=0pt, breakindent=0pt, columns=fullflexible]
You will act as a strict judge. You need to assess whether current knowledge from doc2 is required to accurately answer the given query based on the two provided documents (doc1 and doc2) and the given answer. Doc1 represents known knowledge, while doc2 represents current knowledge. Your task is to determine if the answer relies on doc2 to be accurately provided. Additionally, evaluate whether the answer is short (an entity or judgment) and correct.

If the answer is incorrect or not concise, return false.
If the necessary key information is found in the known knowledge doc1, also return false.
If the answer is a long answer and not of entity type, return false.
If the query is unnatural, not a complete query, or awkwardly phrased, return false.
The answer should result from multi-hop reasoning or multi-step reasoning, where multi-step reasoning indicates that the generated query is challenging and requires reasoning or calculation to answer, and only if the answer is concise should you return true.
You are extremely strict, and any requirements not met should result in a return of false.

Please return the result in JSON format as follows:
{
    "reason": "Evaluate against the above conditions step by step, considering whether the query and answer meet the conditions. Use English to justify, and if they do, list the sections from doc2 that assist in answering the query.",
    "decision": "true/false"
}

\end{lstlisting}
\end{tcolorbox}

\subsection{Details Prompts for Agents}
\label{app:prompt_agent}
The prompts for the \textbf{Expoloer Agent \symbolexplorer} and \textbf{Critic Agent \symbolecritic} are shown below.

\begin{tcolorbox}[
enhanced jigsaw,
breakable,
pad at break*=1mm,
colback=white!95!gray,
colframe=gray!50!black,
title={Prompts for WebWalker}]
\small
\textbf{The Expoloer Agent \symbolexplorer}
\begin{lstlisting}[breaklines=true, xleftmargin=0pt, breakindent=0pt, columns=fullflexible]
Digging through the buttons to find quailty sources and the right information. You have access to the following tools:

(*@\textcolor{blue}{\{tool\_descs\}}@*)

Use the following format:

Question: the input question you must answer
Thought: you should always think about what to do
Action: the action to take, should be one of [{tool_names}]
Action Input: the input to the action
Observation: the result of the action
... (this Thought/Action/Action Input/Observation can be repeated zero or more times)

Begin!

(*@\textcolor{blue}{\{query\}}@*)
\end{lstlisting}
\textbf{The Critic Agent \symbolecritic}
\begin{lstlisting}[breaklines=true, xleftmargin=0pt, breakindent=0pt, columns=fullflexible]
(*@\textbf{Critic} @*)
You are a critic agent. Your task is to analyze the given observation and extract information relevant to the current query. You need to decide if the observation contains useful information for the query. If it does, return a JSON object with a "usefulness" value of true and an "information" field with the relevant details. If not, return a JSON object with a "usefulness" value of false.
**Input:**
    - Query: "<Query>"
    - Observation: "<Current Observation>"
**Output (JSON):**
{
  "usefulness": true,
  "information": "<Extracted Useful Information>"
}
Or, if the observation does not contain useful information:
{
  "usefulness": false
}
- Query: (*@\textcolor{blue}{\{Query\}}@*)
- Observation: (*@\textcolor{blue}{\{Observation\}}@*)
(*@\textbf{Answer} @*)
You are a critic agent. Your task is to evaluate whether the accumulated useful information is sufficient to answer the current query. If it is sufficient, return a JSON object with a "judge" value of true and an "answer" field with the answer.
If the information is insufficient, return a JSON object with a "judge" value of false.
**Input:**
    - Query: "<Query>"
    - Accumulated Information: "<Accumulated Useful Information>"
**Output (JSON):**
{
    "judge": true,
    "answer": "<Generated Answer>"
}
Or, if the information is insufficient to answer the query:
{
    "judge": false
}
- Query: (*@\textcolor{blue}{\{Query\}}@*)
- Accumulated Information: (*@\textcolor{blue}{\{Information\}}@*)
\end{lstlisting}
\end{tcolorbox}

\section{Details for Evaluation}
\label{app:eval}
\subsection{Evaluator}
\label{app:evaluater}
The evaluator prompt is shown in Figure~\ref{fig:evaluatoer}.
\begin{figure}[t]
\begin{tcolorbox}[
enhanced jigsaw,
breakable,
pad at break*=1mm,
colback=white!95!gray,
colframe=gray!50!black,
title={CoT-QA Evaluator}]
\small
\begin{lstlisting}[breaklines=true, xleftmargin=0pt, breakindent=0pt, columns=fullflexible]
You are a teacher grading a quiz.
You are given a question, the context the question is about, and the student's answer. You are asked to score the student's answer as either CORRECT or INCORRECT, based on the context.
Write out in a step by step manner your reasoning to be sure that your conclusion is correct. Avoid simply stating the correct answer at the outset.

Example Format:
QUESTION: question here
CONTEXT: context the question is about here
STUDENT ANSWER: student's answer here
EXPLANATION: step by step reasoning here
GRADE: CORRECT or INCORRECT here

Grade the student answers based ONLY on their factual accuracy. Ignore differences in punctuation and phrasing between the student answer and true answer. It is OK if the student answer contains more information than the true answer, as long as it does not contain any conflicting statements. Begin! 


QUESTION: (*@\textcolor{blue}{\{{query}\}}@*)
CONTEXT:  (*@\textcolor{blue}{\{{answer}\}}@*)
STUDENT ANSWER: (*@\textcolor{blue}{\{{result}\}}@*)
EXPLANATION
GRADE:"""
\end{lstlisting}
\end{tcolorbox}
\caption{The prompt for evaluation.}
\label{fig:evaluatoer}
\end{figure}

\section{Case Study}
\subsection{Reasoning Error}
\label{sec:reasonerror}
As shown in Table~\ref{table:reason}, this question requires first locating the webpage related to the \textcolor{blue}{\textbf{Inclusive Connections Lounge}}, followed by a comprehensive understanding of the information on the page to calculate the required time.
In such cases, it is also necessary to account for the system's ability to perform time calculations or reasoning. 
Consequently, even when the source page is successfully located, errors might still occur if the system fails to process the time correctly.
\subsection{Time Cut-off}
\label{sec:time_cutoff}
As shown in Table~\ref{table:cut_off}, the cutoff date for o1's temporal data is October 2023, rendering it unable to provide answers regarding web information published beyond this point.

\begin{table*}[t]
\small
\centering
\begin{tabularx}{0.89\textwidth}{cX}
\toprule
\textbf{Root Url} & \url{https://www.mrs.org/}\\
\midrule
\textbf{Question}  &  How many hours in total would a person spend if they attended the \textcolor{blue}{\textbf{Inclusive Connections Lounge}} activities from December 1 to 6, 2024, at the MRS Fall Meeting?
\\
\midrule
\textbf{Answer}  & 66 hours \\
\midrule
\textbf{Source Website} & \url{https://www.mrs.org/meetings-events/annual-meetings/2024-mrs-fall-meeting/meeting-events/broadening-participation/inclusive-connections-lounge} \\
\midrule
\textbf{Website Information} & \includegraphics[width=7cm]{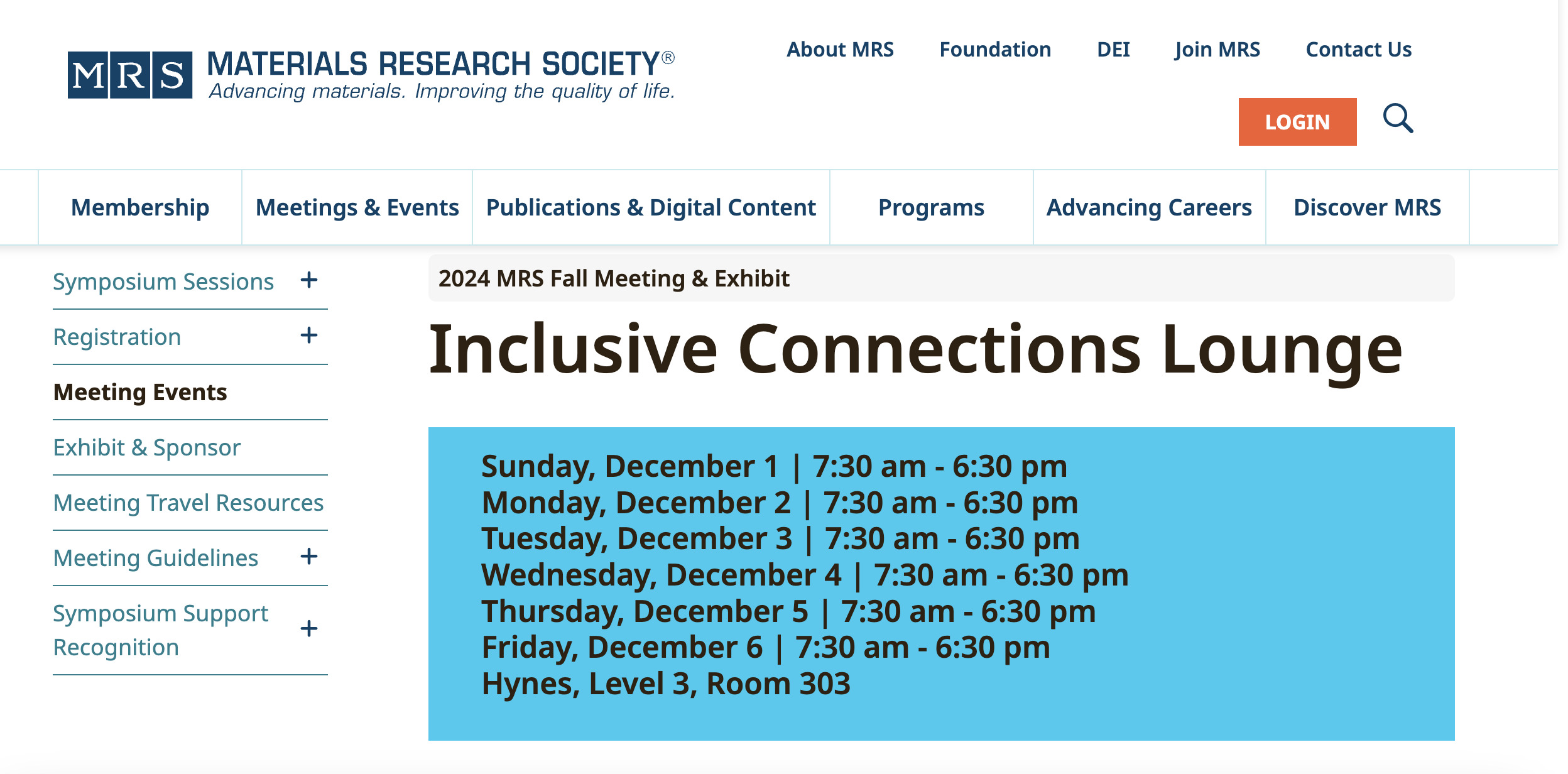} \\
\bottomrule 
\end{tabularx}
\caption{The case requiring reasoning capability in web traversal task.}
\label{table:reason}
\end{table*}

\begin{table}[t]
\small
\centering
\begin{tabularx}{0.49\textwidth}{cX}
\toprule
\textbf{Question}  & Where and when will the \textbf{2025} MRS Fall  Meeting take place? \\
\midrule
\textbf{Answer}  & Boston, Massachusetts;  November 30 to December 5, 2025. \\
\midrule
\textbf{Prediction} & As of my knowledge cutoff in \textcolor{blue}{\textbf{October 2023}}, the MRS has not yet announced the exact dates or location for the 2025 MRS Fall Meeting. \\
\bottomrule 
\end{tabularx}
\caption{The case of time cutoff in predictions generated by o1.}
\label{table:cut_off}
\end{table}

\end{document}